\documentclass[lettersize,journal]{IEEEtran}
\usepackage{amsmath,amsfonts}
\usepackage{algorithmicx}
\usepackage{algorithm}
\usepackage{array}
\usepackage[caption=false,font=normalsize,labelfont=sf,textfont=sf]{subfig}
\usepackage{textcomp}
\usepackage{stfloats}
\usepackage{url}
\usepackage{verbatim}
\usepackage{enumitem}
\usepackage{graphicx}
\usepackage{cite}
\usepackage{booktabs}
\usepackage{amssymb}
\usepackage{multirow}
\usepackage{csquotes}
\usepackage{paralist}
\usepackage{threeparttable}
\usepackage{colortbl}

\newcommand{\equref}[1]{Eq.~\eqref{#1}}

\newcommand{\tabref}[1]{Table~\ref{#1}}

\newcommand\eg{\emph{e.g.}} 
\newcommand\ie{\emph{i.e.}}

\newcommand\etc{\emph{etc. }}
 
\newcommand\etal{\emph{et al.}}

\usepackage{algpseudocode}
\usepackage[pagebackref,breaklinks,colorlinks]{hyperref}
\begin{document}

\title{Contrastive Conditional Latent Diffusion for Audio-visual Segmentation}

\author{Yuxin Mao,~
Jing Zhang*,~
Mochu Xiang,~
Yunqiu Lv,~
Dong Li,~
Yiran Zhong,~
Yuchao Dai*\\
\IEEEcompsocitemizethanks{\IEEEcompsocthanksitem Yuxin Mao, Jing Zhang, Mochu Xiang, Yunqiu Lv, and Yuchao Dai are with School of Electronics and Information, Northwestern Polytechnical University, and Shaanxi Key Laboratory of Information Acquisition and Processing, Xi'an, China.
% \IEEEcompsocthanksitem Jing Zhang is with School of Computing, the Australian National University, Canberra, Australia.
\IEEEcompsocthanksitem Dong Li and Yiran Zhong are with Shanghai AI Laboratory, China.
\IEEEcompsocthanksitem
% Y. Lv and J. Zhang contributed equally. J. Zhang (zjnwpu@gmail.com) and 
Jing Zhang (zjnwpu@gmail.com) and Yuchao Dai (daiyuchao@nwpu.edu.cn) are the corresponding authors. 
\IEEEcompsocthanksitem This research was supported in part by National Natural Science Foundation of China (62271410, 12150007) and by the Fundamental Research Funds for the Central Universities.
Yuxin Mao is sponsored by the Innovation Foundation for Doctoral Dissertation of Northwestern Polytechnical University (CX2024014).

% \IEEEcompsocthanksitem The source code and experimental results are publicly available via our project page: \url{xx}.
}
}

\newcommand{\toreviewer}[1]{\vspace{0.1em}\noindent \textcolor{blue}{\textbf{#1 \hspace{0.1em}}}}
\def\Fix#1{{\color{black}{{#1}}}}
\def\Fixtwo#1{{\color{black}{{#1}}}}

\markboth{Journal of \LaTeX\ Class Files,~Vol.~14, No.~8, August~2021}%
{Shell \MakeLowercase{\textit{et al.}}: A Sample Article Using IEEEtran.cls for IEEE Journals}


\maketitle

\begin{abstract}
% The task of AVS 
Audio-visual Segmentation (AVS) is conceptualized as a conditional generation task, where audio is considered as the conditional variable for segmenting the sound producer(s). In this case, audio should be extensively explored to maximize its contribution for the final segmentation task. We propose a contrastive conditional latent diffusion model for audio-visual segmentation (AVS) to thoroughly investigate the impact of audio, where the correlation between audio and the final segmentation map is modeled to guarantee the strong correlation between them.
To achieve semantic-correlated representation learning, our framework incorporates a latent diffusion model. 
The diffusion model learns the conditional generation process of the ground-truth segmentation map, resulting in ground-truth aware inference during the denoising process at the test stage. As our model is conditional, it is vital to ensure that the conditional variable contributes to the model output.
We thus extensively model the contribution of the audio signal by minimizing the density ratio between the conditional probability of the multimodal data, \eg~conditioned on the audio-visual data, and that of the unimodal data, \eg~conditioned on the audio data only.
In this way, our latent diffusion model via density ratio optimization explicitly maximizes the contribution of audio for AVS, which can then be achieved with contrastive learning as a constraint, where the diffusion part serves as the main objective to achieve maximum likelihood estimation, and the density ratio optimization part imposes the constraint. 
By adopting this latent diffusion model via contrastive learning, we effectively enhance the contribution of audio for AVS. The effectiveness of our solution is validated through experimental results on the benchmark dataset.
Code and results are online via our project page: \url{https://github.com/OpenNLPLab/DiffusionAVS}.

% We propose a latent diffusion model with contrastive learning for audio-visual segmentation (AVS) to thoroughly investigate the impact of audio. 
% The task of AVS is conceptualized as conditional generation, where audio is considered the conditional variable for segmenting sound producer(s). The correlation between audio and the final segmentation map is modeled to ensure its contribution to this new approach.
% To achieve semantic-correlated representation learning, our framework incorporates a latent diffusion model. This diffusion model learns the conditional generation process of the ground-truth segmentation map, resulting in ground-truth aware inference during the denoising process at the test stage. As our model is conditional, it is vital to ensure that the conditional variable contributes to the model output.
% Furthermore, contrastive learning is incorporated into our framework to capture audio-visual correspondence. This approach consistently maximizes the mutual information between model predictions and audio data.
% By adopting this latent diffusion model via contrastive learning, we effectively enhance the contribution of audio for AVS. The effectiveness of our solution is validated through experimental results on the benchmark dataset.
% Code and results are online via our project page: \url{https://github.com/OpenNLPLab/DiffusionAVS}.

\end{abstract}


\section{Introduction}
\label{sec:intro}

\IEEEPARstart{A}{udio-visual} segmentation (AVS)~\cite{zhou_AVSBench_ECCV_2022,hao_aaai_2024_avsbg,li_catr_acmmm_2023,liu_avs_acmmm_2023,mao_iccv_2023_ecmvae} aims to accurately segment the region in the image that produces the sound from the audio. 
Unlike semantic segmentation~\cite{chen_deeplab_pami_2017} or instance segmentation~\cite{he2017mask,wan2025instance}, AVS involves identifying the foreground object(s) responsible for producing the given sound in the audio. 
Due to the usage of multimodal data, \ie~audio and visual, \Fix{AVS typically relies on multimodal learning, where various fusion strategies are explored to integrate audio and visual data.}
Most of these methods rely on the cross modality attention layer~\cite{zhou_AVSBench_ECCV_2022} or the transformer module~\cite{liu_avs_acmmm_2023, li_catr_acmmm_2023} to implicitly fuse the audio-visual feature.
% However, such methods rely purely on fitting the discrete samples in the dataset.

\Fix{We argue that without using audio as guidance, the visual information alone is insufficient for training the AVS model through regression-based learning.}
This \enquote{guided} attribute also distinguishes AVS from other multimodal binary segmentation, \ie~RGB-Depth salient object detection~\cite{ucnet_sal}, where each unimodal data can achieve reasonable prediction. 
With the above understanding of AVS, we find it essential to ensure the audio contribution for AVS, or the model output should be correlated with the audio. In this paper, we aim to extensively explore the contribution of audio for AVS with better data alignment modeling.

Specifically, we define the task of AVS as a conditional generation task, which aims to extensively explore the correlation between audio-visual input (the conditional variable) and the segmentation of the sound producer(s) (target).
Conditional generation can be achieved via maximizing the conditional log-likelihood with likelihood based generative models, \ie~conditional variational auto-encoders (CVAE)~\cite{structure_output,kingma2013auto}, diffusion models~\cite{diffusion_model_raw,ho_ddpm_NIPS_2020},~\etc
Building upon the CVAE, Mao~\etal~\cite{mao_iccv_2023_ecmvae} propose to maximize the likelihood via an evidence lower bound (ELBO) with a latent space factorization strategy, proving its general effectiveness. 
This approach demonstrates the utility of employing a generative model to represent a meaningful multimodal latent space and its effectiveness in enhancing the performance of AVS.
However, the latent space in CVAE contains less semantically related information, and it suffers from the posterior collapse issue~\cite{lucas2019understanding}. 
On the other hand, diffusion models are proven more effective in producing semantic correlated latent space~\cite{label_efficient_diffusion}. 
Therefore, we introduce the diffusion model to our AVS task to ensure the extraction of semantic information from the conditional variable.
In particular, we encode the ground-truth segmentation map and use it as the target of the diffusion model, which is destroyed and generated by the diffusion model via the forward and denoising process. Furthermore, we encode the audio-visual pair and use it as the condition, leading to a conditional generative process.

% Given the strong mapping ability from visual to the segmentation map~\cite{chen2017rethinking}, directly using diffusion models with the multimodal data, \eg~audio and visual, has a risk of achieving conditional likelihood maximization without effective modeling of the audio, especially with the less diverse training dataset~\cite{zhou_AVSBench_ECCV_2022} for AVS.

Based on the conditional diffusion modeling, we argue that besides the maximization of the multimodal conditional generation, extra constraints should be introduced, such that the model output is well-aligned with the audio signal.
The alignment is achieved via minimizing the density ratio $r(\mathbf{y},\mathbf{x}^v,\mathbf{x}^a)$ between the conditional probability of the multimodal data $p(\mathbf{y}|\mathbf{x}^v,\mathbf{x}^a)$ and the unimodal data $p(\mathbf{y}|\mathbf{x}^a)$, where $\mathbf{x}^v$ and $\mathbf{x}^a$ represent the visual and audio data respectively, and $\mathbf{y}$ is the segmentation map.
Additionally, $p(\mathbf{y}|\mathbf{x}^v,\mathbf{x}^a)$ is conditioned on the audio-visual data $\mathbf{x}^v$ and $\mathbf{x}^a$ respectively, while $p(\mathbf{y}|\mathbf{x}^a)$ is conditioned only on the audio data $\mathbf{x}^a$.
In this context, $\mathbf{y}$ represents the desired segmentation map indicating the sound producer(s).
Further, we claim that minimizing the density ratio can be achieved through contrastive learning.

Contrastive learning, which is initially introduced for metric learning~\cite{chopra2005learning,dimension_reduction_lecun}, serves the purpose of acquiring a discriminative feature representation. 
In the context of representation learning, contrastive loss~\cite{oord2018representation} is employed to ensure that positive samples outputted by the network are maximally similar, while negative samples are distinctly dissimilar. 
Traditionally, in the unimodal setting~\cite{wang2021dense,o2020unsupervised,chaitanya2020contrastive,xie2021detco}, data augmentation is utilized to construct positive/negative pairs.
However, for our specific multimodal task, we construct positive/negative samples based on paired/unpaired audio-visual latent variables. 
Subsequently, the contrastive learning solution is derived from a density ratio perspective, enhancing the contribution and semantic richness of the audio guidance. 
We establish the necessity of aligning the audio signal with the prediction by minimizing a density ratio, and contrastive learning emerges as an effective approach to imposing such alignment constraints.

% With the conditional latent diffusion model and contrastive learning based on density ratio minimizing strategy, our model can achieve more effective latent space modeling and better exploration of audio for AVS.
\Fix{Our conditional latent diffusion model, coupled with contrastive learning via density ratio minimization, effectively models latent space and enhances audio exploration for AVS.}
% the contribution of audio modes.
Extensive experimental results demonstrate that our proposed pipeline achieves state-of-the-art AVS performance, especially on the more challenging multiple sound source segmentation dataset.

We summarize our main contributions as:
\begin{compactitem}
    \item We rethink audio-visual segmentation (AVS) as a supervised conditional generation task, to explore the semantic relationship between the guiding input (audio) and the resulting output (segmentation maps).
    \item  We introduce the latent diffusion model, and the maximum likelihood estimation objective to guarantee the ground-truth aware inference.
    \item  A density ratio is introduced to impose the alignment constraint between audio and model output via contrastive learning to maximize the contribution of audio for the desired output within our latent diffusion model.
    \item Experimental results demonstrate that our proposed method achieves state-of-the-art segmentation performance. Extensive ablation experiments further validate the effectiveness of each component in our approach.

    % \item We rethink AVS as a guided conditional generation task, aiming to extensively explore the semantic correlation between the guidance (the audio) and the final output (the segmentation maps).
    % \item  We introduce the latent diffusion model, and its maximum likelihood estimation objective guarantees
    % the ground-truth aware inference.
    % \item  We adopt contrastive learning to our framework to ensure the distinction of the audio representation to achieve an effective latent diffusion model.
    % \item Experiments show that our proposed method achieves state-of-the-art segmentation performance and extensive ablation experiments demonstrate the effectiveness of each component.
\end{compactitem}
\section{Related Work}
\noindent\textbf{Audio-Visual Segmentation.}
Audio-visual segmentation (AVS) is a challenging, newly proposed problem that predicts pixel-wise masks for the sound producer(s) in a video sequence given audio information. 
To tackle this issue, Zhou~\etal~\cite{zhou_AVSBench_ECCV_2022} propose an audio-visual
segmentation benchmark and provide pixel-level annotations. The dataset contains five-second videos and audio, and the binary mask is used to indicate the pixels of sounding objects for the corresponding audio.
Subsequently, they present a simple baseline, an encoder-decoder network based on temporal pixel-wise audio-visual interaction.
Building upon this work, CATR~\cite{li_catr_acmmm_2023} introduces a comprehensive approach that incorporates both spatial and temporal dependencies in an audio-visual combination.
CMMS~\cite{liu_avs_acmmm_2023} extends the AVS tasks to the instance level.
Hao~\etal~\cite{hao_aaai_2024_avsbg} present an audio-visual correlation module with a bidirectional generation consistency module to ensure audio-visual signal consistency.
However, this fusion strategy only considers correlations at the feature level and does not capture the intrinsic characteristic of AVS, namely, the guiding role of audio.
Considering the role of audio as guidance for guided
% uniqueness of this task is that audio serves as guidance, leading to guided 
multimodal binary segmentation, Mao~\etal~\cite{mao_iccv_2023_ecmvae} employ a multimodal VAE with latent space factorization to model the distribution of audio and visual, aiming to maximize the contribution of audio for AVS.

% Due to its binary segmentation nature, models for salient object detection (SOD) and video foreground segmentation (VOS)~\cite{mahadevan_3DC_VOS_2020,duke_sstvos_cvpr_2021,zhang_ebm_sod_nips_2021,mao_transformerSOD_2021} (segmenting the foreground attracts human attention) are usually treated as baselines. However, the uniqueness of this task is that audio serves as guidance, leading to guided multimodal binary segmentation.

\noindent\textbf{Diffusion Models for Segmentation.}
Diffusion model~\cite{diffusion_model_raw,ho_ddpm_NIPS_2020, denoising_diffuion,mao2024tavgbench,li2024tri} is the most popular image-generation approach aiming to learn data distribution through the iterative forward noise-adding process and the reverse denoising process. In recent days, researchers have found that it is also an effective representative learning method to capture essential features or structures~\cite{diffusion_representation_learning,Preechakul_2022_CVPR,traub2022representation,kingma2021on,label_efficient_diffusion,zhu_CDCD_ICLR_2023}. For image segmentation,
% many works have studied how diffusion model extract strong features and help improve the performance. In  
\cite{baranchuk_label_efficient_DDPMSeg_2021}
% it first 
demonstrates that the feature representation learned by a pre-trained diffusion model can significantly benefit zero-shot image segmentation. Pix2Seq-D~\cite{chen2022generalist} extend the bit-diffusion~\cite{chen2022analog} for panoptic segmentation.
\cite{asiedu_DecoderPretrain_arxiv_2022} propose a decoder pre-training strategy to pre-train the decoder of the diffusion UNet for image segmentation.
\cite{lai2023ddps} use the diffusion model for the mask prior modeling.
\cite{segdiff_image_segmentation_with_diffusion, Rahman_2023_CVPR} utilize diffusion probabilistic model for medical image segmentation. 
Most of the mentioned works only study unimodal image segmentation. However, our work investigates representative features across multiple modalities and semantic connections between them.

\noindent\textbf{Contrastive Learning for Representation Learning.}
Contrastive loss~\cite{chopra2005learning,dimension_reduction_lecun,chen2020simple} is introduced for distance metric learning to decide whether the pair of samples is similar or dissimilar.
% which takes pair of examples ($\mathbf{x}$ and $\mathbf{x}'$) as input and train the network ($E$) to predict whether they are similar (from the same class: $\mathbf{y}_\mathbf{x}=\mathbf{y}_{\mathbf{x}'}$) or dissimilar (from different classes: $\mathbf{y}_\mathbf{x}\neq \mathbf{y}_{\mathbf{x}'}$). 
% Taking a step further, triplet loss~\cite{Distance_Metric_Learning,large_scale_online_learning,facenet} achieves distance metric learning by using triplets, including a query sample ($\mathbf{x}$), it is a positive sample ($\mathbf{x}^{+}$) and a negative sample ($\mathbf{x}^{-}$). The goal of triplet loss is to push the difference of similarity between positive and negative samples to the query sample to be greater than a predefined margin parameter. 
Taking a step further, triplet loss~\cite{Distance_Metric_Learning,large_scale_online_learning,facenet} uses triplets to push the difference of similarity between positive ($\mathbf{x}^{+}$) and negative samples ($\mathbf{x}^{-}$) to the query sample ($\mathbf{x}$) to be greater than a predefined threshold and achieves better feature representation learning.
% However, the unbalanced problem will limit the performance due to the fact that it only learns from one negative samples while ignoring other negative samples.  
% achieves distance metric learning by using triplets, including a query sample ($\mathbf{x}$), it is a positive sample ($\mathbf{x}^{+}$) and a negative sample ($\mathbf{x}^{-}$). The goal of triplet loss is to push the difference of similarity between positive and negative samples to the query sample to be greater than a predefined margin parameter.
% By pulling similar concepts to be closer in the embedding space and pushing the dissimilar ones to be far apart, triplet loss achieves better feature representation learning.
% However, one of the main issues is that it only learns from one negative sample, ignoring the dissimilarity with all the other candidate negative samples, leading to unbalanced metric learning. 
% To solve this problem, 
Later,~\cite{npair_loss} introduces N-pair loss to learn from multiple negative samples.  
% for balanced metric learning.
% Contrastive loss, triplet loss, and N-pair loss produce one positive sample for the query sample $\mathbf{x}$, 
% SimCLR~\cite{chen2020simple} produces two noise versions of $\mathbf{x}$ via different data augmentation strategies, and then maximizes agreement between differently augmented views of the same sample via a contrastive loss in latent space. 
% Consider a sample $\mathbf{x}$ and its two correlated views $\tilde{\mathbf{x}}_i$ and $\tilde{\mathbf{x}}_j$, 
% % where $\tilde{\mathbf{x}}_i$ and $\tilde{\mathbf{x}}_j$ is defined 
% termed as the positive pair, \cite{chen2020simple} design a self-supervised feature representation framework with a base encoder $f(\cdot)$ (\eg~ResNet backbone) and a projection head $g(\cdot)$ (\eg~a MLP), where the former is for image backbone feature extraction, and the latter produces latent space representation, where contrastive loss is applied. 
% In this case, the contrastive prediction task aims to identify $\tilde{\mathbf{x}}_j$ from $\{\tilde{\mathbf{x}}_k\}_{k\neq i}$ for a given $\tilde{\mathbf{x}}_i$. 
The main strategy to achieve self-supervised contrastive learning is constructing positive/negative pairs via data augmentation~\cite{xie2021propagate,li2021dense,wang2021dense,van2021unsupervised,o2020unsupervised,chaitanya2020contrastive,xie2021detco}.
% Prototypical contrastive learning (PCL)~\cite{li2020prototypical,li2020prototypical,lin2022prototypical,Prototypical_Graph_Contrastive_Learning,Prototypical_Momentum_Contrastive_Learning,du2022weakly,mo2022siamese,Yue_2021_CVPR,toering2022self} aims to bridge contrastive learning with clustering, where prototypes are introduced as latent variables to find the maximum-likelihood estimation of model parameters. Specifically, prototypical contrastive learning optimize both instance discrimination and semantic structure similarity, where semantic structure of the data is also encoded into the embedding space for more fine-grained representation learning. 
Instead of model instance/image discrimination, dense contrastive learning~\cite{Wang_2022_CVPR_SimSet,wang2021exploring}, widely used in segmentation tasks, aims to explore pixel-level similarity. 
% \cite{wang2021exploring} introduces pixel-wise contrastive learning
% and applies it 
% to semantic segmentation. 
Specifically, 
% the positive/negative pairs, namely 
memory bank~\cite{wang2020xbm,chen2020improved,he2020momentum,chen2020simple,wang2022contrastive} stores historical samples from the same image or different images to make up the positive/negative pool, thereby improving the discriminative capabilities of the model. In this paper, we explore contrastive learning to extensively maximize the alignment of model output with the audio signal from a density ratio perspective, leading to both effective guided segmentation and distinction of our solution with existing techniques~\cite{sun2023learning}.

\noindent\textbf{Uniqueness of Our Solutions.}
Although diffusion models have been explored in segmentation tasks, our method aims to use a conditional latent diffusion model to learn an effective multimodal latent space.
Within the representation learning method, we learn an effective representation of the ground-truth segmentation maps, which is subsequently used to support in the segmentation results.
Instead of employing the diffusion model directly in a separate pipeline, we propose a strategy that utilizes contrastive learning to minimize the audio density ratio. This strategy imposes an explicit constraint on the latent space of the diffusion model and allows us to maximize the contribution of audio for localizing the sound source to achieve high quality segmentation.
\begin{figure*}[!htp]
\begin{center}
\includegraphics[width=0.90\linewidth]{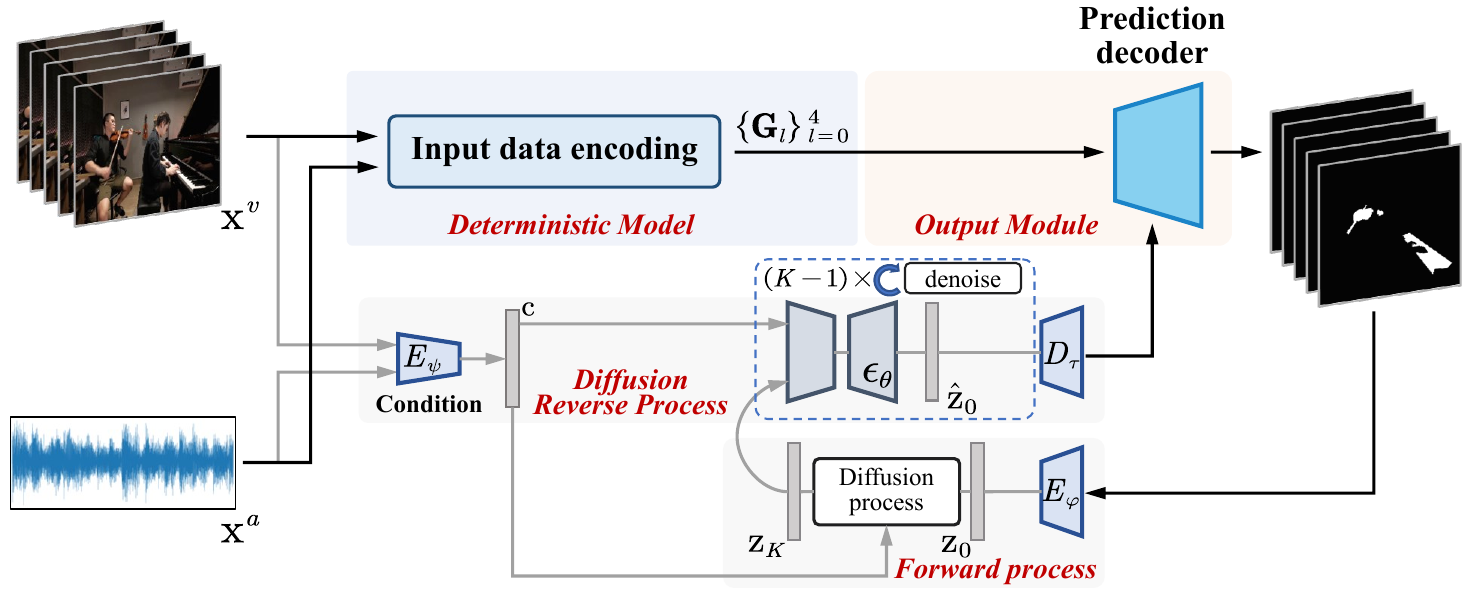}
\end{center}
% \vspace{-4.0mm}
\caption{\textbf{Overview of the proposed method for audio-visual segmentation.} 
It contains three main processes: 
1) a deterministic model to perform input data encoding with multi-scale deterministic audio-visual features ($\{\mathbf{G}_l\}_{l=0}^4$); 
2) a conditional latent diffusion model is used to provide semantic meaningful latent representation, where contrastive learning is ignored for clear presentation. Note that the forward process with ground truth encoding ($E_\phi$) is only used during training;
3) a prediction decoder to aggregate latent representation and deterministic features for the final segmentation map.
}
% \vspace{-4.0mm}
\label{fig:model_overview}
\end{figure*}

\section{Method}
Given the training dataset $D\!=\!\{\mathbf{X}_i,\mathbf{y}_i\}_{i=1}^N$ with the input data $\mathbf{X}\!=\!\{\mathbf{x}^v,\mathbf{x}^a\}$ 
($\mathbf{x}^v$ represents the input video with continuous frames~\cite{zhou_AVSBench_ECCV_2022}, $\mathbf{x}^a$ is the audio of the current clip) 
and ground-truth segmentation map $\mathbf{y}$, the goal of AVS is to segment the sound producer(s) from $\mathbf{x}^v$ with the guidance from $\mathbf{x}^a$.
$i$ indexes the samples, which are omitted for clear presentation. 
As discussed in Sec.~\ref{sec:intro}, AVS is unique in that audio serves as guidance to achieve guided binary segmentation, making it different from conventional multimodal learning~\cite{baltruvsaitis2018multimodal}, where each modality contributes nearly equally to the final output. 
Given this distinction, we define AVS as a conditional generation task, where our objective is to maximize the likelihood of the conditional distribution $p(\mathbf{y}|\mathbf{x}^v,\mathbf{x}^a)$.

% The overview of our proposed method can be seen in Fig.~\ref{fig:model_overview}.
% (see Fig.~\ref{fig:model_overview} for an overview
% % whole pipeline 
% of our method).
% Let $D=\{X_i,y_i\}_{i=1}^N$ to be the training dataset with $i$ as the index. $X=\{\{x^v_t\}_{t=1}^T,x^a\}$ denotes the input data, \ien~the visual $\{x^v_t\}_{t=1}^T$ for $T$ continuous frames, audio $x^a$ of the current clip. $y=\{y_t\}_{t=1}^T$ are the ground-truth segmentation map, \ien~the segmentation maps (we omit $t$ for clear presentation).
% The whole pipeline of our method is illustrated in Fig.~\ref{fig:model_overview}.

% We aim to segment objects that are producing the sound $x^a$ in a video $\{x^v_t\}_{t=1}^T$. 

We resort to diffusion models for our AVS task (see Sec.~\ref{subsec_conditional_latent_diffusion}), aiming to model the distribution of $p(\mathbf{y}|\mathbf{x}^v,\mathbf{x}^a)$.
Further, considering the constraint that the segmentation map should be well-aligned with the audio signal, we introduce contrastive learning (see Sec.~\ref{subsec_contrastive_learning}) as a constraint to our framework.
We use the constraint to explicitly model the correspondence between visual and audio latent variables to guarantee the effectiveness of the conditional variables. 
Finally, we present our pipeline and detailed implementation details of each module in Sec.~\ref{subsec_objective_function}.
The overview of the proposed method is shown in Fig.~\ref{fig:model_overview}.

% We present a latent diffusion model via contrastive learning to extensively explore the contribution of audio for effective audio-visual segmentation (see Fig.~\ref{fig:model_overview} for an overview of the proposed method).
% Toward this goal, we model the latent code over the segmentation maps under the audio-visual pair as a condition using the latent diffusion model, then feed the learned latent distribution into the decoder to achieve the segmentation. 
% Further, we perform contrastive learning under the positive latent code (encoded from the paired audio-visual data as the condition) and the negative latent code (encoded from the unpaired audio-visual data as the condition) to exploit the \enquote{paired information} between visual and audio inputs.

% Our network is composed of five main modules:
% 1) Latent Encoder that maps the ground-truth into a low dimensional latent code; 2) Latent Conditional Diffusion Model to learn a meaningful latent space; 3) Audio-Visual network to produce deterministic feature maps; 4) PredictionNet that employs stochastic features and deterministic features to produce the final segmentation results; 5) P/N pair construction and contrastive learning.
% We will introduce each module as follows.

\subsection{Conditional Latent Diffusion Model for AVS}
\label{subsec_conditional_latent_diffusion}
% With a conditional latent diffusion model, we model the conditional distribution $p(\mathbf{y}|\mathbf{x}^v,\mathbf{x}^a)$, where a latent diffusion model is learned to estimate the conditional ground-truth density function, achieving ground-truth aware inference.
We model the conditional distribution $p(\mathbf{y}|\mathbf{x}^v,\mathbf{x}^a)$ using a conditional latent diffusion model. 
Specifically, the latent diffusion model \Fix{learns} to estimate the conditional ground-truth density function, achieving ground-truth aware inference.

% Our network is composed of five main modules:
% 1) Latent Encoder that maps the ground-truth into a low dimensional latent code; 2) Latent Conditional Diffusion Model to learn a meaningful latent space; 3) Audio-Visual network to produce deterministic feature maps; 4) PredictionNet that employs stochastic features and deterministic features to produce the final segmentation results; 5) P/N pair construction and contrastive learning.
% We will introduce each module as follows.

\noindent\textbf{Latent Space Modeling.}
We develop two encoders to encode the ground-truth segmentation map and the audio-visual input signal, respectively,
% into a latent space, 
where the former is designed to achieve ground-truth aware inference, and the latter is to achieve the projection from input space to feature space.

We denote $E_{\varphi}$ as the ground-truth encoder to encode the ground-truth segmentation map, denoted by $\mathbf{z}_0$. 
Specifically, we have $\mathbf{z}_0=E_{\varphi}(\mathbf{y})\in \mathbb{R}^{B\times D}$, where $B$ represents the batch size and $D$ corresponds to the latent space dimension. 
It is worth mentioning that our approach for encoding the ground-truth is similar to the posterior computation strategy in ECMVAE~\cite{mao_iccv_2023_ecmvae}. 
However, a key distinction lies in that we explicitly model the latent variable using a diffusion model, allowing it to follow any distribution. 
In contrast, ECMVAE relies on assuming a Gaussian distribution for the latent variable utilizing the re-parameterization trick~\cite{kingma2013auto}.
To construct the ground-truth latent encoder $E_{\varphi}$, we employ a structure comprising five convolutional layers, followed by leakyReLU and batch normalization. 
The output channels for these layers are [16, 32, 64, 64, 64], respectively. 
Subsequently, we utilize two fully connected layers to generate a latent code of size $D=24$.

Moreover, we define the conditional input encoder as $E_{\psi}$, the encoder takes audio-visual pairs as input and outputs a conditional latent variable $\mathbf{c}$.
Thus, we obtain $\mathbf{c}=E_{\psi}(\mathbf{x}^v, \mathbf{x}^a)$.
In order to encode audio-visual signals simultaneously, $E_{\psi}$ is divided into two branches, namely the visual branch and the audio branch. 
The visual branch consists of five convolutional layers and two fully connected layers, which share the same $E_{\varphi}$ structure. The audio branch involves two fully connected layers. Further, the visual and audio features are concatenated along the channel dimension, and another two fully connected layers are used to get the final conditional embedding $\mathbf{c}$.

\begin{figure}[t!]
\begin{center}
\includegraphics[width=0.98\linewidth]{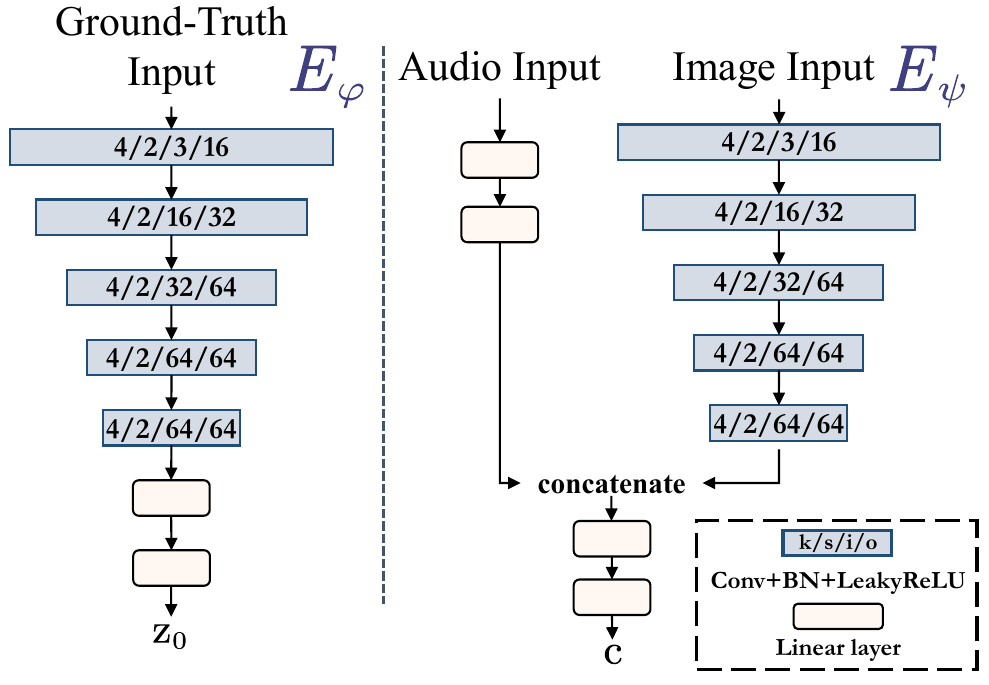}
\end{center}
% \vspace{-4.0mm}
\caption{\textbf{Detailed structure of the latent encoders}, where \enquote{k/s/i/o} indicates the kernel size, stride, in channel, and out channel.}
% \vspace{-4.0mm}
\label{fig:latent_encoder}
\end{figure}

For ease of understanding, the more detailed structure of latent encoders $E_{\varphi},E_{\psi}$ is shown in Fig.~\ref{fig:latent_encoder}.
It should be noted that our chosen latent encoders are lightweight enough and do not impose additional computational overhead. 
In the experimental section, we will present a detailed analysis of the model's parameter complexity and computational efficiency.

\noindent\textbf{Conditional Latent Diffusion Model.}
Given the latent code $\mathbf{z_0}$, our conditional latent diffusion model aims to learn its distribution to restore the ground-truth information during testing. 
Firstly, we review latent diffusion models~\cite{diffusion_model_raw,ho_ddpm_NIPS_2020,rombach_stable_diffusion_cvpr_2022}. 
Then, we present our conditional diffusion model, which gradually diffuses $\mathbf{z}_0$ to $\mathbf{z}_K\sim\mathcal{N}(0,\mathbf{I})$, and restores $\mathbf{z}_0$ back from $\mathbf{z}_K$ under $\mathbf{c}$ as conditional.

% with the audio-visual data as a conditional variable. 

\noindent\textit{\textbf{Latent Diffusion model.}}
The latent diffusion model is built upon a generative Markov chain, which converts a simple known distribution, (\eg~a Gaussian) into a target distribution. 
The fundamental concept behind the diffusion model~\cite{diffusion_model_raw,ho_ddpm_NIPS_2020} involves the deliberate and gradual degradation of a latent code's structure through an iterative forward diffusion process. 
Subsequently, the reverse diffusion process is employed to reconstitute structures within the sample.

Following the standard diffusion procedure, the initial latent data representation $\mathbf{z}_0$ undergoes a gradual transformation into an analytically tractable distribution, denoted as $\pi(\mathbf{z})=\mathcal{N}(0,\mathbf{I})$. 
This conversion occurs through iterative application of a Markov diffusion kernel $T_\pi(\mathbf{z}|\mathbf{z}';\beta)$, utilizing a diffusion rate parameter $\beta$, as expressed by:

\begin{equation}
    q(\mathbf{z}_k|\mathbf{z}_{k-1})=T_\pi(\mathbf{z}_k|\mathbf{z}_{k-1};\beta_k).
\end{equation}
% and the final analytically tractable distribution $\pi(z)$
% % , \ie~$\mathcal{N}(0,\mathbf{I})$,
% is defined as:
% \begin{equation}
%     \pi(z) = \int T_\mathcal{N}(\mathbf{z}|\mathbf{z}';\beta) \pi(\mathbf{z}') d\mathbf{z}'.
% \end{equation}
The forward trajectory of the diffusion model is thus:
\begin{equation}
q(\mathbf{z}_{0,...,K})=q(\mathbf{z}_0)\prod_{k=1}^K q(\mathbf{z}_k|\mathbf{z}_{k-1}),
\end{equation}
where the diffusion kernel $q(\mathbf{z}_k|\mathbf{z}_{k-1})$ is defined as Gaussian in~\cite{diffusion_model_raw,ho_ddpm_NIPS_2020}
% either Gaussian diffusion 
with an identity-covariance:
\begin{equation}
    q(\mathbf{z}_k|\mathbf{z}_{k-1}) = \mathcal{N}(\mathbf{z}_k;\sqrt{1-\beta_k}\mathbf{z}_{k-1},\beta_k\mathbf{I}).
\end{equation}
A notable property of the forward diffusion process is that it admits sampling $\mathbf{z}_k$ at arbitrary timestep $k$ in closed form:
\begin{equation}
\label{eq_diffusion_process}
    q(\mathbf{z}_k|\mathbf{z}_0)=\mathcal{N}(\mathbf{z}_k;\sqrt{\bar{\alpha}_k}\,\mathbf{z}_0,(1-\bar{\alpha}_k)\mathbf{I}),
\end{equation}
where $\alpha_k=1-\beta_k$ and $\bar{\alpha}_k=\prod_{s=1}^k\alpha_s$. \equref{eq_diffusion_process} explains the stochastic diffusion process, where no learnable parameters are needed, and a pre-defined set of hyper-parameters $\{\beta\}_{k=1}^K$ will lead to a set of latent variables $\{\mathbf{z}\}_{k=1}^K$.

The generative process or the denoising process is then to 
% The generative distribution is trained to 
% restore the data distribution \Mochu{restore the sample} 
restore the sample via:
\begin{equation}
p_\theta(\mathbf{z}_{0,...,K})=p(\mathbf{z}_{K})\prod_{k=1}^K p_\theta(\mathbf{z}_{k-1}|\mathbf{z}_{k}),
\end{equation}
where $p(\mathbf{z}_{K})=\pi(\mathbf{z})=\mathcal{N}(0,\mathbf{I})$ in our case.
For Gaussian diffusion, during learning, only the mean ($\mu$) and variance ($\Sigma$)
% for a Gaussian diffusion kernel 
are needed to be estimated, leading to:
\begin{equation}
    p_\theta(\mathbf{z}_{k-1}|\mathbf{z}_{k})=\mathcal{N}(\mathbf{z}_{k-1};\mu_\theta(\mathbf{z}_{k},k),\Sigma_\theta(\mathbf{z}_{k},k)),
\end{equation}
where $\theta$ represents model parameters. $\Sigma$ is set as hyper-parameters by~\cite{ho_ddpm_NIPS_2020}. Specifically, $\Sigma_\theta(\mathbf{z}_{k},k)=\beta_k\mathbf{I}$ is used for stable training, which means only $\mu_\theta(\mathbf{z}_{k},k)$ is learned.

\noindent\textit{\textbf{Conditional diffusion model for AVS.}}
For our AVS task, with the ground-truth latent encoder $\mathbf{z}_0=E_{\varphi}(\mathbf{y})$, \equref{eq_diffusion_process} provides the diffusion process by gradually destroying $\mathbf{z}_0$ to obtain $\mathbf{z}_K\sim\mathcal{N}(0,\mathbf{I})$. Our conditional generation process aims to restore $\mathbf{z}_0$ given the input conditional variable $\mathbf{c}=E_{\psi}(\mathbf{x}^v, \mathbf{x}^a)$, where $\mathbf{c}$ is the feature embedding of our audio-visual input, leading to the conditional generative process $p_\theta(\mathbf{z}_{k-1}|\mathbf{z}_{k},\mathbf{c})$.
In our implementation, we concatenate the conditional variable $\mathbf{c}$ with the noisy ground-truth latent variable $\mathbf{z}_K$, to achieve conditional generation.
It is important to note that since the binary ground truth lacks appearance information, we utilize the audio-visual fused feature $\mathbf{c}$ instead of the only audio feature to correlate the \enquote{visual} sound producer(s) with the audio data.
% as no appearance information in encoded in the output feature $\mathbf{z}_0$. Our goal is to learn semantic rich latent space, where we aim to correlated \enquote{visual} sound producer(s) with audio, and the fused information in our case can provide \enquote{visual} guidance. 
We thus sample from $p_\theta(\mathbf{z}_0|\mathbf{c})$ via:
\begin{equation}
\begin{aligned}
    \label{eq_conditional_generation}
    &p_\theta(\mathbf{z}_0|\mathbf{c})=\int p_\theta(\mathbf{z}_{0,...,K}|\mathbf{c}) \text{d}\mathbf{z}_{1,...,K},\\
&p_\theta(\mathbf{z}_{0,...,K}|\mathbf{c})=p(\mathbf{z}_K)\prod_{k=1}^K p_\theta(\mathbf{z}_{k-1}|\mathbf{z}_k,\mathbf{c}).
\end{aligned}
\end{equation}

Following the simplified diffusion model objective~\cite{ho_ddpm_NIPS_2020}, with the re-parameterization trick~\cite{kingma2013auto}, a noise estimator $\epsilon_\theta$ is designed to regress the actual noise $\epsilon$ added to $\mathbf{z}_k$ via:
% Specifically, 
% % To train the denoising UNet,
% we adopt the simplified objective proposed by Ho \etal~\cite{ho_ddpm_NIPS_2020} to train our latent conditional diffusion model.
% of  and an encoder $E_{\psi}$ to get the audio-visual embedding as a condition. The latent code and the condition can be represented as:
% \begin{equation}
%     \begin{aligned}
%     \mathbf{z_0}=E_{\varphi}(\mathbf{y}), \quad\text{and}\quad \mathbf{c}=E_{\psi}(x^a,x^v)
%     \end{aligned}
% \end{equation}
\begin{equation}
    \begin{aligned}
    \label{ddpm_loss}
    \mathcal{L}_{\text{diffusion}}(\theta):=\mathbb{E}_{\mathbf{z},\mathbf{c}, \epsilon\sim \mathcal{N}(0,\mathbf{I}), k}\left[\left\|\epsilon-\epsilon_\theta\left(\mathbf{z}_k, \mathbf{c}, k\right)\right\|^2\right].
    \end{aligned}
\end{equation}

The forward and reverse process of our proposed conditional latent diffusion model can be shown in Fig.~\ref{fig:model_overview}. 
In the training phase, one-step denoising is completed through sampling with a randomly sampled timestep $t$.
And the denoising objective is under the supervision of \equref{ddpm_loss}.
At inference time, given the conditional latent variable $\mathbf{c}$ of the audio-visual pair and random noise $\mathbf{z}_K\!\sim\!\mathcal{N}(0,\mathbf{I})$, our model samples $p_\theta(\mathbf{z}_0|\mathbf{c})$ via \equref{eq_conditional_generation} by gradually performing denoising.
% \MYX{describe how to perform condition here.}
% SDFusion: Multimodal 3D Shape Completion, Reconstruction, and Generation

\noindent\textit{\textbf{The structure of $\epsilon_\theta$.}}
As described above, the noise estimator $\epsilon_\theta$ constitutes the central component within the diffusion model.
Following the conventional practice in designing the diffusion models~\cite{ho_ddpm_NIPS_2020}, $\epsilon_\theta$ can be designed as a \enquote{encoder-decoder} structure.
In our implementation, we design eight fully connected layers followed by leakyReLU activation to ensure lightweight. 
The former four layers are \enquote{encoder}, and the latter four layers are \enquote{decoder}.

\subsection{Contrastive Representation Learning}\label{subsec_contrastive_learning}
In the context of the conditional diffusion process described in \equref{eq_conditional_generation}, the efficacy of $p_\theta(\mathbf{z}_0|\mathbf{c})$ holds considerable significance.
In our multimodal scenario, the proficiency of $p_\theta(\mathbf{z}_0|\mathbf{c})$ relies on the representational quality of the multimodal conditional variable $\mathbf{c}$, where the guidance of audio data facilitates the segmentation of visual data.
% The uniqueness of AVS lies in the strong requirement for audio, as it provides guidance to achieve the so-called guided segmentation. 
The distinctiveness of AVS stems from its significant reliance on audio, as it serves as a guiding force to accomplish guided segmentation.
However, without extra constraint, the audio feature representation~\cite{hershey_VGGish_icassp_2017} could become dominated by the visual modality, leading to a less effective representation of audio, which is critical for AVS.

\noindent\textbf{Density Ratio Modeling.} 
We initiate the process by considering the conditional probability $p(\mathbf{y}|\mathbf{x}^v,\mathbf{x}^a)$ and proceed to derive the density ratio. 
This ratio acts as a constraint, focusing on maximizing the contribution of the audio signal. 
Employing Bayes' rule, we obtain:

\begin{equation}
    \begin{aligned}
p(\mathbf{y}|\mathbf{x}^v,\mathbf{x}^a)=\frac{p(\mathbf{y},\mathbf{x}^v,\mathbf{x}^a)}{p(\mathbf{x}^v,\mathbf{x}^a)}=\frac{p(\mathbf{x}^v|\mathbf{y},\mathbf{x}^a)p(\mathbf{y}|\mathbf{x}^a)}{p(\mathbf{x}^v|\mathbf{x}^a)}.
    \end{aligned}
\end{equation}
We define the density ratio $r(\mathbf{y},\mathbf{x}^v,\mathbf{x}^a)$ as: 
\begin{equation}
\label{eq_likelihood_ratio}
   % r(\mathbf{y},\mathbf{x}^v,\mathbf{x}^a)=\frac{p(\mathbf{y}|\mathbf{x}^v,\mathbf{x}^a)}{p(\mathbf{y}|\mathbf{x}^a)}= \frac{p(\mathbf{x}^v|\mathbf{y},\mathbf{x}^a)}{p(\mathbf{x}^v|\mathbf{x}^a)}.
   r(\mathbf{y},\mathbf{x}^v,\mathbf{x}^a)= \frac{p(\mathbf{x}^v|\mathbf{y},\mathbf{x}^a)}{p(\mathbf{x}^v|\mathbf{x}^a)}=\frac{p(\mathbf{y}|\mathbf{x}^v,\mathbf{x}^a)}{p(\mathbf{y}|\mathbf{x}^a)}.
\end{equation}
To maximize the contribution of the audio data, our objective is to minimize the density ratio $r(\mathbf{y},\mathbf{x}^v,\mathbf{x}^a)$, thereby avoiding poor alignment between the output $\mathbf{y}$ and the audio data $\mathbf{x}^a$.

% Given the correlation between the density ratio with the objective function of contrastive learning, we claim that minimizing the density ratio can be achieved via contrastive learning.
% Recall that contrastive learning~\cite{chopra2005learning} aims to maximize the distance of the given sample to its negative sample(s) and minimize its distance to the positive sample.
% Considering the alignment requirement of $\mathbf{y}$ and $\mathbf{x}^a$, the goal of the optimization of $r(\mathbf{y},\mathbf{x}^v,\mathbf{x}^a)$ is then to maximizing $p(\mathbf{y}|\mathbf{x}^a)$ for the paired $\mathbf{y}$ and $\mathbf{x}^a$, and minimizing it otherwise.

Given the correlation between the density ratio and the objective function in contrastive learning~\cite{chopra2005learning,li2023mutual}, we claim that minimizing the density ratio can be attained through contrastive learning methods.
Recall that contrastive learning aims to maximize the distance between a given sample and its negative samples, while simultaneously minimizing its distance to the positive samples.
In light of the alignment requirement between $\mathbf{y}$ and $\mathbf{x}^a$, the primary objective in optimizing $r(\mathbf{y},\mathbf{x}^v,\mathbf{x}^a)$ is to maximize $p(\mathbf{y}|\mathbf{x}^a)$ for the matched pairs of $\mathbf{y}$ and $\mathbf{x}^a$, and minimize it otherwise.

\noindent\textbf{Contrastive Learning to Optimize the Density Ratio.}
As a conditional generative model, we argue that the representativeness of the conditional variable in the latent diffusion model plays an important role in the sample quality, especially for our specific multimodal task, where audio data serves as guidance for the visual data to achieve guided segmentation. 
We will first introduce our conditional variable generation process, \ie~$\mathbf{c}=E_{\psi}(\mathbf{x}^v, \mathbf{x}^a)$, and then 
% We want to have the conditional variable $\mathbf{c}$ to be discriminative enough to explain the alignment of audio and visual. We thus introduce contrastive representation learning to our framework. We will first introduce the conditional feature generation process
% % encoding 
% and then 
present our positive/negative pairs construction for contrastive learning. Our objective is to learn an appropriate distance function, such that the paired audio-visual sound producer(s) data remains in close proximity in the latent space compared to the unpaired data.
This can be achieved via maximizing $p(\mathbf{y}|\mathbf{x}^a)$ for paired samples and minimizing $p(\mathbf{y}|\mathbf{x}^a)$ for unpaired samples.
% \Jing{I'm here!}

% Besides the latent code $\mathbf{z}_0$, we have three variables involved in our framework, namely video $\mathbf{x}^v$, audio $\mathbf{x}^a$, and ground-truth segmentation map $\mathbf{y}$. We design a diffusion model to diffuse and restore the ground-truth information, thus we can sample from $p_\theta(\mathbf{z}_0|\mathbf{c})$ via \equref{eq_conditional_generation}. In that case, we claim that the conditional variable $\mathbf{c}$ should be discriminative enough to distinguish $\mathbf{z}_0$. 
% In other words, given $\mathbf{c}$, the corresponding $\mathbf{z}_0$ should lead to a larger score than  $\mathbf{z}'_0$ of another sound producer(s).

% Based on the above analysis, we present contrastive learning to investigate the contribution of the audio data. 

We claim the conditional variable $\mathbf{c}$ should be discriminative enough to distinguish $\mathbf{z}_0$. 
In other words, given $\mathbf{c}$, the corresponding $\mathbf{z}_0$ should lead to a larger score than  $\mathbf{z}'_0$ of another sound producer(s).
% Specifically, with the audio-visual conditional feature $\mathbf{c}=E_{\psi}(\mathbf{x}^v_i, \mathbf{x}^a_i)$, we define its ground-truth encoding $\mathbf{z}_0=E_{\varphi}(\mathbf{y}_i)$ as its positive sample, and $\mathbf{y}'$ other than $\mathbf{y}_i$ in the mini-batch as the negative samples.
More specifically, utilizing the audio-visual conditional feature $\mathbf{c}=E_{\psi}(\mathbf{x}^v_i, \mathbf{x}^a_i)$, we define its ground-truth encoding $\mathbf{z}_0=E_{\varphi}(\mathbf{y}_i)$ as the positive sample, while considering $\mathbf{y}'$ (distinct from $\mathbf{y}_i$) within the mini-batch as the negative samples.
With the above positive/negative samples, we obtain our contrastive loss as:

\begin{equation}
    \begin{aligned}
    \label{contrastive_loss}
    \mathcal{L}_{\text{contrastive}}=-\mathbb{E}_\mathbf{\mathbf{z}_0}\left[\log \frac{f(\mathbf{z}_0,\mathbf{c})/\tau}{\sum_{\mathbf{z}_0'\in\{\mathcal{N},\mathbf{z}_0\}}f(\mathbf{z}'_0,\mathbf{c})/\tau}\right],
    \end{aligned}
\end{equation}
where $\mathbf{z}_0$ is always paired with $\mathbf{c}$, and $\mathcal{N}$ represents the negative samples within the mini-batch, which includes all the samples except $\mathbf{z}_0$.
$f(\mathbf{z}_0,\mathbf{c})=\exp(s(\mathbf{z}_0,\mathbf{c}))$ is the scoring function with $s(\cdot,\cdot)$ as the cosine similarity.
$\tau$ is a temperature parameter and we set $\tau\!=\!1$ in all experiments.

With the utilization of the contrastive loss described in \equref{contrastive_loss}, our objective is to maximize $f(\mathbf{z}_0,\mathbf{c})$. 
This maximization aligns with the goal of enhancing the mutual information between $\mathbf{z}_0$ and $\mathbf{c}$, or equivalently, maximizing $p(\mathbf{y}|\mathbf{x}^a)$ as indicated in \equref{eq_likelihood_ratio} for the paired data.

\subsection{Model Prediction Generation and Training}
\label{subsec_objective_function}
In Sec.~\ref{subsec_conditional_latent_diffusion}, we present our conditional latent diffusion model for learning a conditional distribution $p_\theta(\mathbf{z}_0|\mathbf{c})$ and restoring the ground truth information $\hat{\mathbf{z}}_0$ during inference. Moreover, as described in Sec.~\ref{subsec_contrastive_learning}, the discriminativeness of the conditional variable and its contribution to the final output is constrained via the contrastive learning pipeline. 
As shown in Fig.~\ref{fig:model_overview}, the restored $\hat{\mathbf{z}}_0$ and the input data encoding are fed to the prediction decoder to generate our final prediction.

% given the  
% Our whole framework in Fig.~\ref{fig:model_overview} is composed of a fused feature extraction module for AVS, where TPAVI is used for cross-level multimodal fusion following~\cite{zhou_AVSBench_ECCV_2022}, a contrastive learning module for representative feature generation, a conditional diffusion module to explicitly model the conditional generation process $p(y|x^a,x^v)$, and a prediction decoder to generate our final segmentation map.

\noindent\textbf{Input Data Encoding.}
\Fix{We design a two-branch Audio-Visual network to produce multi-scale deterministic feature maps from the input audio-visual pairs, following the established paradigm of processing each modality through specialized encoders before fusion. 
Similar to ECMVAE~\cite{mao_iccv_2023_ecmvae}, we encode the deterministic audio and visual features through separate branches to leverage modality-specific pre-trained models.
For the audio branch, we preprocess the audio waveform into a spectrogram via short-time Fourier transform and feed it to a frozen VGGish~\cite{hershey_VGGish_icassp_2017} model, which is pre-trained on the large-scale AudioSet~\cite{gemmeke_audioset_icassp_2017} dataset. This specialized audio encoder yields rich audio representations $\mathbf{A}\in \mathbb{R}^{T\times d}$, where $d=128$ is the feature dimension.
For the visual branch, given the video sequence $\mathbf{x}^v$, we extract visual features using either the ImageNet pre-trained ResNet50 backbone~\cite{he_resnet_cvpr_2016} or the PVTv2 backbone~\cite{wang_Pvtv2_CVM_2022}. These visual-specific encoders produce multi-scale visual features denoted as $\mathbf{F}_l\in \mathbb{R}^{T\times c_l\times h_l\times w_l}$, where $c_l$ represents the number of channels, and $(h_l,w_l)=(H,W)/2^{l+1}$. The spatial dimension of the input video is $(H,W)$, and the feature levels are $l\in [1,4]$.
For the ResNet50 backbone, the channel sizes of the four stages are $c_{1:4}=[256, 512, 1024, 2048]$, while for the PVTv2 backbone, they are $c_{1:4}=[64, 128, 320, 512]$.
We further process the visual features $\mathbf{F}_l$ using four convolutional neck modules to obtain $\mathbf{V}_l\in \mathbb{R}^{T\times c\times h_l\times w_l}$, where $c\!=\!128$.
After obtaining the modality-specific representations, we perform multimodal fusion using the temporal pixel-wise audio-visual interaction module~\cite{zhou_AVSBench_ECCV_2022}. This cross-modality attention mechanism explores the correlation between audio features $\mathbf{A}$ and visual features $\mathbf{V}_l$, effectively integrating information from both modalities. Through this fusion process, we obtain the deterministic feature maps $\mathbf{G}_l\in \mathbb{R}^{T\times c\times h_l\times w_l}$ that encode rich audio-visual correlations, forming the foundation for our conditional generation approach to audio-visual segmentation.}

\Fix{This separate encoding followed by fusion approach offers several key advantages. 
First, it allows us to leverage powerful pre-trained models that have been optimized on large-scale datasets specific to each modality, extracting higher-quality modality-specific features. 
Second, the specialized encoders preserve the unique statistical properties and information structures of the audio and visual data before integration. Third, this architecture facilitates more controlled and interpretable cross-modal interaction, as the fusion module can explicitly model how audio cues should guide visual segmentation. Finally, this design aligns well with our conditional generation framework, where audio serves as a guiding condition for the segmentation process, enabling a more precise modeling of the audio-visual relationship.}

\noindent\textbf{Prediction Decoder.}
Since the deterministic features $\mathbf{G}_l$ and stochastic representation $\mathbf{\hat{z}}_0$ are with different feature sizes, to fuse the two items, we perform a latent code expanding module $D_\tau$, which contains one $3\times 3$ convolutional layer, to achieve feature expanding of $\mathbf{\hat{z}}_0$. Specifically, we first expand $\mathbf{\hat{z}}_0$ to a 2D tensor and tile it to the same spatial size as $\mathbf{G}_4$. We define the new 2D feature map as $\mathbf{\hat{z}^\mathbf{e}}_0$. Given that the spatial size of $\mathbf{\hat{z}^\mathbf{e}}_0$ and $\mathbf{G}_4$ are the same, we perform cascaded channel-wise feature concatenation and one $3\times 3$ convolution to obtain $\mathbf{\hat{G}}_4$, which is the same size as $\mathbf{G}_4$.
\Fix{Following the classic work in AVS~\cite{zhou_AVSBench_ECCV_2022}, we adopt Panoptic-FPN~\cite{kirillov_panopticfpn_cvpr_2019} as our decoder to process the mixed features  $\{\mathbf{G}_l \mid l = 1, 2, 3\} \cup \{\mathbf{\hat{G}}_4\}$. This architecture efficiently combines a bottom-up pathway with a top-down pathway featuring lateral connections, creating a feature pyramid that effectively preserves both visual spatial details and multimodal semantic information. The lightweight segmentation head then processes these multi-scale features to generate the final output $\mathbf{M}\in \mathbb{R}^{T\times 1\times H\times W}$. Since our task is binary segmentation (foreground vs. background), we apply the \texttt{sigmoid} activation function to the output, naturally mapping network predictions to probability values between 0 and 1, which is mathematically appropriate for our binary classification objective.}

\noindent\textbf{Objective Function.} As a segmentation task, our model is trained with a cross-entropy loss with the ground-truth segmentation map as supervision. We also have a conditional latent diffusion module and a contrastive learning objective involved, leading to our final objective as:
\begin{equation}
    \label{eq_objective_function}
    \mathcal{L}=\mathcal{L}_{\text{seg}}+\lambda_1 \mathcal{L}_{\text{diffusion}}+\lambda_2 \mathcal{L}_{\text{contrastive}},
\end{equation}
where $\lambda_1$ and $\lambda_2$ are used to balance the two objectives, which are set empirically as 1 and 0.1, respectively.
All proposed modules can be completed through end-to-end training, eliminating the need for additional pre-training.
% : $\{\lambda_1, \lambda_2\}=\{1, 0.1\}$.
61\section{Experimental Results}
\subsection{Setup}
\noindent\textbf{Datasets.}
We utilize the AVSBench dataset~\cite{zhou_AVSBench_ECCV_2022}, which consists of 5,356 audio-video pairs with pixel-wise annotations.
% for conducting our experiments. 
Each audio-video pair in the dataset spans 5 seconds, and we trim the video to include five consecutive frames by extracting the video frame at the end of each second. 
The AVSBench dataset is further divided into two subsets: semi-supervised Single Sound Source Segmentation (S4), where only the first frame is labeled, and fully supervised Multiple Sound Source Segmentation (MS3), where all frames are labeled. 
The S4 subset contains 4,922 videos, while the MS3 subset contains 424 videos. 
For training and testing, we follow the conventional splitting from the AVSBench dataset~\cite{zhou_AVSBench_ECCV_2022} and perform training and testing with S4 and MS3, respectively.
% and We train our model using the training set and evaluate its performance on the testing set.

\noindent\textbf{Evaluation Metrics.}
We assess the audio-visual segmentation performance using the same evaluation metrics as AVSBench~\cite{zhou_AVSBench_ECCV_2022}, namely Mean Intersection over Union (mIoU) and F-score. 
The F-score is formulated as follows: $F_{\beta}=\frac{(1+\beta^2 \times \text{precision} \times \text{recall})}{\beta^2\times \text{precision} + \text{recall}}, \beta^2=0.3$. 
Here, both precision and recall are computed based on a binary segmentation map, which is obtained by applying 256 uniformly distributed binarization thresholds in the range $[0, 255]$.

\noindent\textbf{Compared Methods.} We compare our method with published AVS methods, including AVSBench~\cite{zhou_AVSBench_ECCV_2022}, AVS-BiGen~\cite{hao_aaai_2024_avsbg}, ECMVAE~\cite{mao_iccv_2023_ecmvae}, CATR~\cite{li_catr_acmmm_2023}, CMMS~\cite{liu_avs_acmmm_2023} \Fix{and AVSegFormer~\cite{gao2024avsegformer}}.
To strictly keep accordance with the settings in previous work~\cite{zhou_AVSBench_ECCV_2022}, we also compare the performance with related segmentation tasks, such as video foreground segmentation models (VOS)~\cite{mahadevan_3DC_VOS_2020, duke_sstvos_cvpr_2021}, RGB image based salient object detection models~\cite{mao_transformerSOD_2021, zhang_ebm_sod_nips_2021}.
We set up the comparison due to the binary video segmentation nature of AVS. 
Being consistent with AVSBench, we also use two backbones, ResNet50~\cite{he_resnet_cvpr_2016} and PVT~\cite{wang_Pvtv2_CVM_2022} initialized with ImageNet~\cite{deng_imagenet_cvpr_2009} per-trained weights, to demonstrate that our proposed model achieves consistent performance improvement under different backbones.
\Fixtwo{For a fair comparison, we establish consistent experimental protocols across methods. Regarding CATR, their paper presents two experimental settings: (1) a baseline setting that maintains consistency with the training protocol of AVSBench, and (2) an enhanced setting utilizing additional AOT-enhanced annotations~\cite{yang2021associating_aot}. To ensure fair comparison, we specifically reference their results from the baseline setting. Similarly, AVSegFormer presents two training configurations: (1) a standard setting with $224\times 224$ input resolution that aligns with the configuration of AVSBench, and (2) an enhanced setting with $512\times 512$ resolution that achieves better performance through increased input size. To maintain consistent experimental protocols, we specifically compare with their $224\times 224$ configuration results, ensuring architectural comparisons are conducted under equivalent conditions.}

\noindent\textbf{Implementation Details.}
Our proposed method is trained end-to-end using the Adam optimizer~\cite{Kingma_Adam_ICLR_2015} with default hyper-parameters for 15 and 30 epochs on the S4 and MS3 subsets. The learning rate is set to $10^{-4}$ and the batch size is 4. All the video frames are resized to the shape of $224\times 224$. 
For the latent diffusion model, we use the cosine noise schedule and the noise prediction objective in \equref{ddpm_loss} for all experiments. The diffusion steps $K$ is set as 20. To accelerate sampling, we use the DDIM~\cite{song_DDIM_ICLR_2020} with 10 sampling steps.

\begin{table}[t]
    \caption{\textbf{Quantitative results on the AVSBench dataset} in terms of mIOU and F-score under S4 and MS3 settings. 
    We both report the performance with R50 and PVT as a backbone for the results of comparison methods and Ours.
    \Fix{* denotes that the training datasets are supplemented annotation with AOT~\cite{yang2021associating_aot}.
    For AVSegFormer~\cite{gao2024avsegformer}, we only report the performance when trained at the common $224\times 224$ resolution.}}
    \label{tab:main_results_on_avsbench}
    \centering
    \small
    \setlength{\tabcolsep}{1.0mm}
      \renewcommand{\arraystretch}{1.3}
    {
        \begin{threeparttable}
        \begin{tabular}{cccccc}
        \toprule[1.1pt]
        & \multirow{2}{*}{Methods} & \multicolumn{2}{c}{S4}                      & \multicolumn{2}{c}{MS3}                     \\
        \cmidrule(r){3-4}  \cmidrule(r){5-6}
                     &         & mIoU                 & F-score              & mIoU                 & F-score   \\
        \midrule
        \multirow{2}{*}{VOS}& 3DC~\cite{mahadevan_3DC_VOS_2020}    & 57.10   & 0.759   & 36.92   & 0.503    \\
        & SST~\cite{duke_sstvos_cvpr_2021}                        & 66.29   & 0.801   & 42.57   & 0.572    \\
        \midrule
        \multirow{2}{*}{SOD}& iGAN~\cite{mao_transformerSOD_2021}   & 61.59   & 0.778   & 42.89   & 0.544    \\
        & LGVT~\cite{zhang_ebm_sod_nips_2021}                       & 74.94   & 0.873   & 40.71   & 0.593    \\
        \midrule
        & AVSBench (R50)~\cite{zhou_AVSBench_ECCV_2022}   & 72.79   & 0.848   & 47.88   & 0.578    \\
        & AVSBench (PVT)~\cite{zhou_AVSBench_ECCV_2022}   & 78.74   & 0.879   & 54.00   & 0.645    \\
        & AVS-BiGen (R50)~\cite{hao_aaai_2024_avsbg}   & 74.13   & 0.854   & 44.95   & 0.568 \\
        & AVS-BiGen (PVT)~\cite{hao_aaai_2024_avsbg}   & 81.71   & 0.904   & 55.10   & 0.668 \\
        & ECMVAE (R50)~\cite{mao_iccv_2023_ecmvae}     & 76.33   & 0.865   & 48.69   & 0.607  \\
        \multirow{2}{*}{AVS}& ECMVAE (PVT)~\cite{mao_iccv_2023_ecmvae}     & 81.74   & 0.901   & 57.84   & 0.708  \\
        & CATR (R50)~\cite{li_catr_acmmm_2023}         & 74.8    & 0.866   & 52.8    & 0.653           \\
        & CATR (PVT)~\cite{li_catr_acmmm_2023}         & 81.4    & 0.896   & 59.0    & 0.700  \\
        & \Fix{CATR (R50)*~\cite{li_catr_acmmm_2023}}  & \Fix{74.9}    & \Fix{0.871}   & \Fix{53.1}    & \Fix{0.656}           \\
        & \Fix{CATR (PVT)*~\cite{li_catr_acmmm_2023}}  & \Fix{84.4}    & \Fix{0.913}   & \Fix{62.7}    & \Fix{0.745}  \\
        & CMMS~\cite{liu_avs_acmmm_2023}               & 81.29   & 0.886   & 59.5    & 0.657           \\
        & \Fix{AVSegFormer (R50)~\cite{gao2024avsegformer}}  & \Fix{76.45} & \Fix{0.859} & \Fix{49.53} & \Fix{0.628}           \\
        & \Fix{AVSegFormer (PVT)~\cite{gao2024avsegformer}}  & \Fix{\textbf{82.06}} & \Fix{0.899} & \Fix{58.36} & \Fix{0.693}   \\ 
        & Ours (R50)               & 75.80          & 0.869          & 49.77          & 0.621           \\
        & Ours (PVT)               & 81.51 & \textbf{0.903} & \textbf{59.62} & \textbf{0.712}  \\
        \bottomrule[1.1pt]
        \end{tabular}
        \end{threeparttable}
    }
    % \vspace{-5.0mm}
\end{table}

\begin{figure*}[!htp]
\begin{center}
\includegraphics[width=0.98\linewidth]{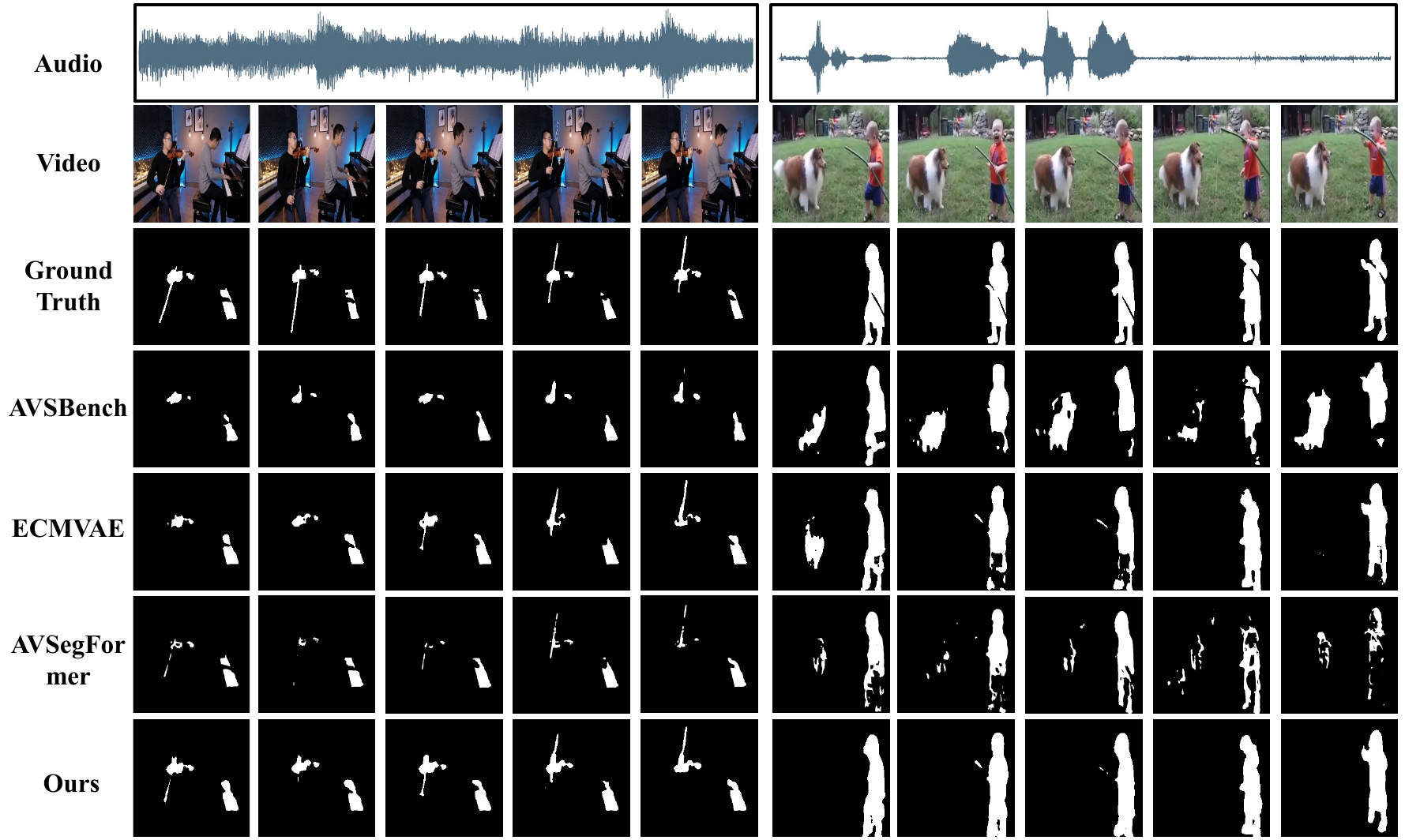}
\end{center}
% \vspace{-4.0mm}
\caption{\textbf{Qualitative comparison with existing method
% on our proposed method and AVSBench~\cite{zhou_AVSBench_ECCV_2022}
} under the fully-supervised MS3 setting. Our proposed method produces much more accurate and high-quality segmentation maps and provides a more accurate sound source localization performance.}
% \vspace{-4.0mm}
\label{fig:main_compare}
\end{figure*}

\subsection{Performance Comparison}
% \Jing{I'm here}
\noindent\textbf{Quantitative Comparison.} Generally, we define our task as a multimodal binary segmentation task, where the input includes both visual and audio, and the output is a binary map showing the sound producer(s). We find a related and similar setting is salient object detection, where the output is also a binary map, localizing the foreground object(s) that attract human attention. In this way, to prepare the comparison methods, we also adapt the existing state-of-the-art (SOTA) salient object detection models to our multimodal binary segmentation task and show the performance of those models in \tabref{tab:main_results_on_avsbench}, where \enquote{VOS} contains video salient object detection models, and \enquote{SOD} lists the SOTA salient object detection models.
Based on the quantitative results obtained from \tabref{tab:main_results_on_avsbench}, we observe that direct adaptation of salient object detection models to AVS fails to achieve reasonable performance. The main reason is that although both salient object detection and AVS are categorized as binary segmentation, the former relies mainly on the visual input, while the latter depends greatly on the audio modality to localize the sound producer(s).

In the \enquote{AVS} section of \tabref{tab:main_results_on_avsbench}, we show performance comparison of various methods and ours on the AVSBench dataset under different settings (S4 and MS3).
\Fix{Our method consistently outperforms state-of-the-art AVS methods on both MS3 and S4 subsets, achieving notable improvements in mIoU (59.62) and F-score (0.712).}
There is a consistent performance improvement of our proposed method compared to CATR~\cite{li_catr_acmmm_2023}, regardless of whether \enquote{R50} or \enquote{PVT} is used as the backbone. In particular, 0.11 and 0.62 higher mIOU than CATR is obtained on the two subsets with the \enquote{PVT} backbone.
Moreover, the performance of our method significantly surpasses that of ECMVAE~\cite{mao_iccv_2023_ecmvae}, an AVS method based on generative models (VAE). This comparison highlights that, despite the fact that ECMVAE employs intricate strategies involving complex multimodal latent space factorization and constraints, its capacity to model the latent space falls short in comparison to our approach utilizing a conditional latent diffusion model.
It is worth noting that our \enquote{R50} based model slightly outperforms the LGVT~\cite{zhang_ebm_sod_nips_2021} under the S4 subset, despite LGVT using a swin transformer~\cite{liu_swin_iccv_2021} backbone, while AVSBench (R50) performs worse than LGVT. 
This suggests that exploring matching relationships between visual objects and sounds is more important than using a better visual backbone for AVS tasks.
\Fixtwo{Notably, our method demonstrates superior performance over AVSegFormer~\cite{gao2024avsegformer} in three out of four metrics across both datasets. This performance advantage stems from our latent diffusion architecture and contrastive loss design, which effectively model the correlation between video and audio modalities, leading to better audio-guided segmentation results. Specifically, on the S4 dataset, while achieving higher F-score due to our strength in sounding object localization, we observe slightly lower mIoU performance. This can be attributed to the single-source characteristic of S4 dataset, where mIoU primarily reflects the refinement of segmentation boundaries rather than the accuracy of sounding object localization, which is relatively straightforward in single-source scenarios.} Despite these achievements, our model maintains a lightweight architecture where $E_{\varphi}$, $E_{\varphi}$, $D_{\tau}$, and $\epsilon_\theta$ collectively contribute only 4M parameters, resulting in a total of 94.48M parameters when incorporating the PVT backbone. \Fixtwo{This parameter count is substantially more efficient compared to AVSegFormer's 186.05M parameters and CATR's 118.38M parameters while achieving better performance.}

\Fixtwo{We further compare the performance of our model with AVSSBench~\cite{zhou2024avss}, CATR~\cite{li_catr_acmmm_2023} and AVSegFormer~\cite{gao2024avsegformer}} on the AVSBench-semantic datasets (AVSS)~\cite{zhou2024avss} dataset. 
\Fixtwo{Compared to AVSegFormer, our model demonstrates consistent improvements with absolute margins of 1.4 and 1.3 in mIoU and F-score metrics, respectively. These performance gains are particularly pronounced on complex datasets containing multiple sounding targets and rich semantic information, as shown in~\tabref{tab:avss-comparison}. This superior performance can be attributed to our model's enhanced capability in modeling audio-visual correlations using the proposed diffusion framework, which becomes more evident when handling sophisticated scenarios with diverse audio sources and semantic contexts.}
The consistent performance across multiple datasets (AVSBench-S4, MS3, and now AVSS) provides substantial evidence for the robustness and adaptability of our approach. This additional experiment reinforces our claim that recasting AVS as a conditional generation task with audio guidance offers a generalizable framework for audio-visual segmentation challenges.

\begin{table}[t]
\caption{Quantitative comparisons on AVSBench-semantic datasets (AVSS)~\cite{zhou2024avss} in terms of mIoU and F-score.}
\label{tab:avss-comparison}
\centering
\small
\setlength{\tabcolsep}{1.5mm}
\renewcommand{\arraystretch}{1.3}
\begin{threeparttable}
\begin{tabular}{ccccc}
\toprule[1.1pt]
Task & Method & Backbone & mIoU & F-score \\
\midrule
\multirow{2}{*}{VOS} & 3DC~\cite{mahadevan_3DC_VOS_2020} & R18 & 17.3 & 0.210 \\
                     & AOT~\cite{yang2021associating_aot} & R50 & 25.4 & 0.310 \\
\midrule
\multirow{4}{*}{AVSS} & AVSSBench~\cite{zhou2024avss} & PVT & 29.8 & 0.352 \\
                      & CATR~\cite{li_catr_acmmm_2023}          & PVT & 32.8 & 0.385 \\
                      & AVSegFormer~\cite{gao2024avsegformer}   & PVT & 36.7 & 0.420 \\
                      & \textbf{Ours}                           & \textbf{PVT} & \textbf{38.1} & \textbf{0.430} \\
\bottomrule[1.1pt]
\end{tabular}
\end{threeparttable}
\end{table}

\noindent\textbf{Qualitative Comparison.}
In Fig.~3, we show the qualitative comparison of our method with
\Fixtwo{AVSBench~\cite{zhou_AVSBench_ECCV_2022}, ECMVAE~\cite{mao_iccv_2023_ecmvae} and AVSegFormer~\cite{gao2024avsegformer}.
Among them, AVSBench is the baseline model, ECMVAE is also a generative AVS model similar to ours. Furthermore, AVSegFormer is the most advanced model.}
The visualization samples in Fig.~\ref{fig:main_compare} are selected from the more challenging MS3 subset.
It can be observed that our method tends to output segmentation results with finer details, \ie~an accurate segmentation of the \emph{bow of the violin} and the \emph{piano-key} in the left sample in Fig.~\ref{fig:main_compare}. 
In addition, our method also has the ability to identify the true sound producer, such as the \emph{boy} in the right sample in Fig.~\ref{fig:main_compare}, indicating a better sound localization capability.
\Fixtwo{Compared to AVSegFormer, which adopts a transformer architecture, our model incorporates audio cues explicitly via a conditional latent diffusion process. This enables more accurate localization of sounding objects, especially in complex scenes. 
As a result, AVSegFormer tends to highlight visually salient regions, whereas our model focuses more accurately on sounding objects.}

% In Fig.~\ref{fig:main_compare}, we show the qualitative comparison of our method with \Fixtwo{AVSBench~\cite{zhou_AVSBench_ECCV_2022}, ECMVAE~\cite{mao_iccv_2023_ecmvae} and AVSegFormer~\cite{gao2024avsegformer}.
% Among them, AVSBench is the baseline model, ECMVAE is also a generative AVS model similar to ours. Furthermore, AVSegFormer is the most advanced model.}
% The visualization samples in Fig.~\ref{fig:main_compare} are selected from the more challenging MS3 subset.
% It can be observed that our method tends to output segmentation results with finer details, \ie~an accurate segmentation of the \emph{bow of the violin} and the \emph{piano-key} in the left sample in Fig.~\ref{fig:main_compare}. 
% In addition, our method also has the ability to identify the true sound producer, such as the \emph{boy} in the right sample in Fig.~\ref{fig:main_compare}, indicating a better sound localization capability. 
% While the compared methods segment the two salient foreground objects, ignoring the audio information as guidance.

% % We illustrate the qualitative results under the challenge MS3 subset in Fig.~\ref{fig:main_compare}. 

\subsection{Ablation Studies}
We conduct ablation studies to analyze the effectiveness of our proposed method. All variations of the experiments are trained with the PVT backbone.

\begin{table}[!htp]
    \caption{\textbf{Ablation on the latent diffusion model.} \enquote{E-D} indicates the deterministic \enquote{encoder-decoder} structure. \enquote{CVAE} denotes using CVAE to generate the latent code. \enquote{LDM} is our proposed latent diffusion model}
    % \vspace{-2.0mm}
    \label{tab:ablation_on_ldm}
    \centering
    \small
    \setlength{\tabcolsep}{2.5mm}
    \renewcommand{\arraystretch}{1.3}
    {
        \begin{threeparttable}
        \begin{tabular}{ccccc}
        \toprule[1.1pt]
        \multirow{2}{*}{Methods} & \multicolumn{2}{c}{S4}    & \multicolumn{2}{c}{MS3}         \\
        \cmidrule(r){2-3}  \cmidrule(r){4-5}
                & mIoU    & F-score & mIoU    & F-score        \\
        \midrule
        E-D      & 78.89      & 0.881      & 54.28      & 0.648       \\
        CVAE     & 79.97      & 0.888      & 55.21      & 0.661       \\
        % \midrule
        LDM (Ours)& \textbf{81.02} & \textbf{0.894} & \textbf{57.67} & \textbf{0.698}  \\
        \bottomrule[1.1pt]
        \end{tabular}
        \end{threeparttable}
    }
    % \vspace{-2.0mm}
\end{table}

\begin{figure}[t!]
\begin{center}
\includegraphics[width=0.98\linewidth]{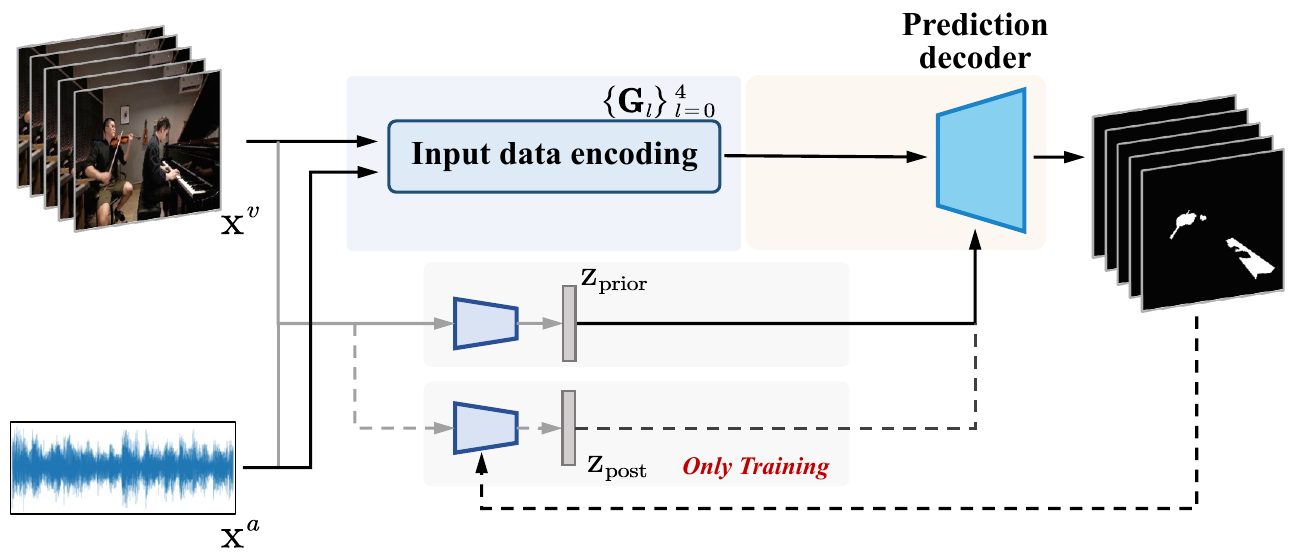}
\end{center}
% \vspace{-4.0mm}
\caption{\textbf{Overview of the CVAE for audio-visual segmentation,} where
% contains prior and posterior latent space, 
the posterior latent code is only used in training.}
% \vspace{-4.0mm}
\label{fig:model_overview_vae}
\end{figure}

\noindent\textbf{Ablation on Latent Diffusion Model.}
As discussed in the introduction section (Sec.~\ref{sec:intro}), a likelihood conditional generative model exactly fits our current conditional generation setting, thus a conditional variational auto-encoder~\cite{structure_output,kingma2013auto} can be a straightforward solution. 
To verify the effectiveness of our latent diffusion model, we design two baselines and show the comparison results
% As the critical component of our proposed method, to verify the impact of the latent diffusion model, we provide two baseline models and show the results
in \tabref{tab:ablation_on_ldm}. Firstly, we design
% : \textbf{1}) 
a deterministic model with a simple encoder-decoder structure (\enquote{E-D}), where the input data encoding $\{\mathbf{G}\}_{l=1}^4$ is feed directly to the prediction decoder (see Fig.~\ref{fig:model_overview}). Note that \enquote{E-D} is the same as AVSBench~\cite{zhou_AVSBench_ECCV_2022}, and we retrain it in our framework and get similar performance as the original numbers reported in their paper. 
Secondly, to explain the superiority of the diffusion model compared with other likelihood based generative models, namely conditional variational auto-encoder~\cite{structure_output} in our scenario, we follow~\cite{ucnet_sal,mao_iccv_2023_ecmvae} and design an AVS model based on CVAE (\enquote{CVAE}).
The full pipeline of the \enquote{CVAE} for the audio-visual segmentation task can be shown in Fig.~\ref{fig:model_overview_vae}.
Note that this structure can be regarded as a simplified version of ECMVAE~\cite{mao_iccv_2023_ecmvae}, which removes the complex multimodal factorization and other latent space constraints.
% \textbf{2}) a conditional variational auto-encoder (\enquote{CVAE}) following~\cite{ucnet_sal}. The structure of \enquote{E-D} is the same as AVSBench~\cite{zhou_AVSBench_ECCV_2022}, we retrain it in our framework and get similar performance as the original report in their paper. 
% CVAE~\cite{structure_output} is introduced to RGB-Depth salient object detection in~\cite{ucnet_sal}, where early fusion is used for latent feature encoding. 
We follow a similar pipeline and perform latent feature encoding based on the fused feature $\{\mathbf{G}_l\}_{l=0}^4$ instead of the early fusion feature due to our audio-visual setting, which is different from the visual-visual setting in~\cite{ucnet_sal}.
Specifically, the CVAE~\cite{structure_output} pipeline for our AVS task consists of an inference process and a generative process, where the inference process infers the latent variable $\mathbf{z}$ by $p_\theta(\mathbf{z}|\mathbf{X})$, and the generative process
% builds a latent variable $\mathbf{z}$ by $p_\theta(\mathbf{z}|\mathbf{X})$ and obtains 
produces the output via $p_\theta(\mathbf{y}|\mathbf{X},\mathbf{z})$.

% \footnote{We will explain the detailed structure of CVAE for the AVS task in the supplementary material.}.
% based generative model. 
Results in~\tabref{tab:ablation_on_ldm} show that generative models can improve the performance of AVS by yielding more meaningful latent space compared with the deterministic models. 
Additionally, the latent diffusion model (LDM) exhibits a more powerful latent space modeling capability than our implemented CVAE counterpart. Note that, as no latent code is involved in \enquote{E-D}, we do not perform contrastive learning. For a fair comparison, the contrastive learning objective $\mathcal{L}_\text{contrastive}$ is not involved in \enquote{CVAE} or \enquote{LDM (Ours)} either.

\begin{table}[t!]
    \caption{\textbf{Ablation on the conditional variable,} where we remove the conditional variable (\enquote{None}), or replace the conditional variable with only audio or visual representation.
    % with  We remove the audio-visual condition, the audio condition, and the visual condition as three comparison variants.
    }
    % \vspace{-2.0mm}
    \label{tab:ablation_on_audio_visual_condition}
    \centering
    \small
    \setlength{\tabcolsep}{2.4mm}
    \renewcommand{\arraystretch}{1.3}
    {
        \begin{threeparttable}
        \begin{tabular}{ccccc}
        \toprule[1.1pt]
        \multirow{2}{*}{Methods} & \multicolumn{2}{c}{S4}    & \multicolumn{2}{c}{MS3}         \\
        \cmidrule(r){2-3}  \cmidrule(r){4-5}
                & mIoU    & F-score & mIoU    & F-score        \\
        \midrule
        None       & 80.04      & 0.889      & 56.12      & 0.671       \\
        Audio      & 80.29      & 0.892      & 56.59      & 0.680       \\
        Visual     & 80.68      & 0.892      & 57.21      & 0.688       \\
        % \midrule
        Audio-Visual (Ours) & \textbf{81.02} & \textbf{0.894} & \textbf{57.67} & \textbf{0.698}  \\
        \bottomrule[1.1pt]
        \end{tabular}
        \end{threeparttable}
    }
    % \vspace{-2.0mm}
\end{table}

\begin{table}[t!]
    \caption{\textbf{Ablation of contrastive learning.} We perform experiments without the $\mathcal{L}_\text{contrastive}$ to show its effectiveness.}
    % \vspace{-2.0mm}
    \label{tab:ablation_on_contrastive_learning}
    \centering
    \small
    \setlength{\tabcolsep}{3.4mm}
    \renewcommand{\arraystretch}{1.3}
    {
        \begin{threeparttable}
        \begin{tabular}{ccccc}
        \toprule[1.1pt]
        \multirow{2}{*}{Methods} & \multicolumn{2}{c}{S4}    & \multicolumn{2}{c}{MS3}         \\
        \cmidrule(r){2-3}  \cmidrule(r){4-5}
                & mIoU    & F-score & mIoU    & F-score        \\
        \midrule
        w/o $\mathcal{L}_\text{contrastive}$    & 81.02 & 0.894 & 57.67 & 0.698  \\
        w   $\mathcal{L}_\text{contrastive}$    & \textbf{81.51} & \textbf{0.903} & \textbf{59.62} & \textbf{0.712} \\
        \bottomrule[1.1pt]
        \end{tabular}
        \end{threeparttable}
    }
    % \vspace{-2.0mm}
\end{table}
\noindent\textbf{Ablation on Audio-Visual Condition.}
To further investigate the effectiveness of the audio-visual conditioning in the training process of the latent diffusion model, we train three models by incorporating different conditional variables $\mathbf{c}$, and present their performance in Table~\ref{tab:ablation_on_audio_visual_condition}. 
Initially, we remove the conditional variable, leading to unconditional generation with $p_\theta(\mathbf{z}_{k-1}|\mathbf{z}_{k})$, which is represented as \enquote{None} in the table. 
Subsequently, we consider unimodal audio or visual as only one conditional variable. 
For this purpose, we simply use the feature of each individual modality before multimodal feature concatenation (refer to $E_\psi$ in Sec.\ref{subsec_conditional_latent_diffusion}), leading to audio/visual as conditional variable based models referred to as \enquote{Audio} and \enquote{Visual} in Table~\ref{tab:ablation_on_audio_visual_condition}.
% by removing the audio condition, the visual condition, and the audio-visual condition in the latent diffusion model. 
% As shown in Table~\ref{tab:ablation_on_audio_visual_condition}, we explore four variants of the different condition types. 
Compared to unconditional generation, conditional generation can provide performance improvements, with the best results achieved when using the audio-visual condition. Furthermore, we can also observe that the performance of using visual data as the conditional variable yields superior performance compared to using audio.
We attribute this observation to two main factors. Firstly, our dataset is small and less diverse, leading to less effective audio information exploration as we pre-trained our model on a large visual image dataset. 
Secondly, the audio encoder is smaller compared with the visual encoder. More investigation will be conducted to address and balance the distribution of data.
In order to ensure a fair comparison, we opted not to perform contrastive learning in the related experiments outlined in Table~\ref{tab:ablation_on_audio_visual_condition}, similar to the ablation on the latent diffusion model.

\noindent\textbf{Ablation on Contrastive Learning.}
We introduce contrastive learning to our framework to learn the discriminative conditional variable $\mathbf{c}$. We then train our model directly without contrastive learning and show its performance as \enquote{w/o $\mathcal{L}_\text{contrastive}$} in Table~\ref{tab:ablation_on_contrastive_learning}, where \enquote{w $\mathcal{L}_\text{contrastive}$} is our final performance in Table~\ref{tab:main_results_on_avsbench}. The improved performance of \enquote{w $\mathcal{L}_\text{contrastive}$} indicates the effectiveness of contrastive learning in our framework.
\begin{table}[t]
    \caption{\textbf{Ablation on the size of the latent space,} where we conduct experiments with different latent sizes.}
    % \vspace{-2.0mm}
    \label{tab:ablation_on_ldm_dimension}
    \centering
    \small
    \setlength{\tabcolsep}{3.8mm}
    \renewcommand{\arraystretch}{1.3}
    {
        \begin{threeparttable}
        \begin{tabular}{ccccc}
        \toprule[1.1pt]
        \multirow{2}{*}{Latent Size} & \multicolumn{2}{c}{S4}    & \multicolumn{2}{c}{MS3}         \\
        \cmidrule(r){2-3}  \cmidrule(r){4-5}
                & mIoU    & F-score & mIoU    & F-score        \\
        \midrule
        $ {D}= 8 $  & 81.04          & 0.892          & 57.28          & 0.689           \\
        $ {D}=16 $  & 81.18          & 0.895          & 57.98          & 0.704           \\
        $ {D}=24 $  & \textbf{81.51} & \textbf{0.903} & \textbf{59.62} & \textbf{0.712}  \\
        $ {D}=32 $  & 80.78          & 0.891          & 57.01          & 0.687           \\
        \bottomrule[1.1pt]
        \end{tabular}
        \end{threeparttable}
    }
    % \vspace{-2.0mm}
\end{table}
Additionally, we observe that contrastive learning performs poorly with the naive encoder-decoder framework, especially with our limited computation configuration, where we cannot construct large enough positive/negative pools. 
However, we find the improvement is insignificant compared to using contrastive learning in other tasks~\cite{han2022expanding}. 
We argue the main reason for this lies in our dataset being less diverse to learn distinctive enough features. 
We will investigate self-supervised learning to further explore the effectiveness of contrastive learning in our framework.
% our model relies on the representation $z$ in this case. Large positive/negative pools can relax the necessity for semantic-correlated representation, as the large sample pools can guarantee the sample-wise distinction.

\begin{table}[t]
\caption{\Fix{\textbf{Ablation on the prediction decoder,} where we conduct experiments under the AVSegFormer architecture.}}
\label{tab:AVSegFormer_diffusion}
\small
\centering
\setlength{\tabcolsep}{1.0mm}
\renewcommand{\arraystretch}{1.3}
\begin{threeparttable}
\begin{tabular}{lcccc}
\toprule[1.1pt]
\multirow{2}{*}{Method} & \multicolumn{2}{c}{S4} & \multicolumn{2}{c}{MS3} \\
\cmidrule(r){2-3}  \cmidrule(r){4-5}
 & mIoU & F-score & mIoU & F-score \\
\midrule
AVSegFormer~\cite{gao2024avsegformer} & 82.06 & 0.899 & 58.36 & 0.693 \\
AVSegFormer~w.~Diffusion (Ours)       & \textbf{82.79} & \textbf{0.910} & \textbf{59.94} & \textbf{0.715} \\
\bottomrule[1.1pt]
\end{tabular}
\end{threeparttable}
\end{table}

\noindent\textbf{Ablation on Size of the Latent Space.} 
We conduct additional ablation experiments to investigate the impact of the latent space size. In the main experiment, we perform parameter tuning and determine that $D = 24$ yields the best results. Here, we proceed to conduct experiments with varied latent sizes and present the performance outcomes in Table~\ref{tab:ablation_on_ldm_dimension}. An obvious observation is that the size of the latent space should not exceed a certain threshold ($D=32$) for the diffusion model, as doing so can lead to significant performance degradation. Conversely, we find that relatively stable predictions are achieved within the latent code dimension range of $D\in [16, 24]$.

\Fix{\noindent\textbf{Ablation on Prediction Decoder.}
We replace the decoder of the model with the transformer decoder in AVSegFormer~\cite{gao2024avsegformer} to demonstrate the applicability of our proposed conditional generation framework under different model frameworks. 
The experimental results are shown in~\tabref{tab:AVSegFormer_diffusion}. 
This demonstrates that our method's contribution extends beyond a specific architecture and represents a general enhancement that can benefit various AVS base models. Note that although alternative decoders such as transformer-based structures (\eg, AVSegFormer) demonstrate strong performance, their higher computational overhead and larger parameter counts motivated us to adopt the more lightweight Panoptic-FPN decoder.}

\subsection{Analysis}
\noindent\textbf{Pre-training Strategy Analysis.} 
As discussed in~\cite{zhou_AVSBench_ECCV_2022}, we also train our model with the full parameters initialized by the weight per-trained on the S4 subset. The performance comparison is shown in \tabref{tab:results_for_pertrain}. 
% The pre-training strategy can facilitate the modeling of audio-visual correspondence by 
It is verified that an effective pre-training strategy is beneficial in all the settings with our proposed method, using \enquote{R50} or \enquote{PVT} as a backbone. We argue the main reason lies in the less diverse and small amount of dataset. In this case, effective transfer learning with suitable model tuning strategies can be a promising research direction to improve the effectiveness of our solution further, \eg~prompt tuning~\cite{lester-etal-2021-power,han2021ptr,li-liang-2021-prefix}.

\begin{table}[t]
    \caption{\textbf{Performance comparison with different initialization strategies} (train from scratch or pre-train on S4) under MS3 setting in terms of mIoU.
    We use the arrows with specific values to indicate the performance gain.
    % the performance mIOU after using the weights pre-trained on the S4 subset.
    }
    % \vspace{-2.0mm}
    \label{tab:results_for_pertrain}
    \centering
    \small
    \setlength{\tabcolsep}{0.8mm}{
        \begin{threeparttable}
        \begin{tabular}{cccc}
        \toprule[1.1pt]
        {Methods} & {From scratch}  &  & {Pre-trained on S4}         \\
        \midrule
        AVSBench (R50)~\cite{zhou_AVSBench_ECCV_2022}   & 47.88 & $\stackrel{+ 6.45}{\longrightarrow}$ & 54.33  \\
        AVSBench (PVT)~\cite{zhou_AVSBench_ECCV_2022}   & 54.00 & $\stackrel{+ 3.34}{\longrightarrow}$ & 57.34  \\
        ECMVAE (R50)~\cite{mao_iccv_2023_ecmvae}   & 48.69 & $\stackrel{+ 8.87}{\longrightarrow}$ & 57.56  \\
        ECMVAE (PVT)~\cite{mao_iccv_2023_ecmvae}   & 57.84 & $\stackrel{+ 2.97}{\longrightarrow}$ & 60.81  \\
        Ours (R50)                             & \textbf{49.77} & $\stackrel{+ 7.82}{\longrightarrow}$ & \textbf{57.59}  \\
        Ours (PVT)                             & \textbf{59.62} & $\stackrel{+ 2.32}{\longrightarrow}$ & \textbf{61.94}  \\
        \bottomrule[1.1pt]
        \end{tabular}
        \end{threeparttable}
    }
    % \vspace{-2.0mm}
\end{table}

\begin{figure}[t]
\begin{center}
\includegraphics[width=0.98\linewidth]{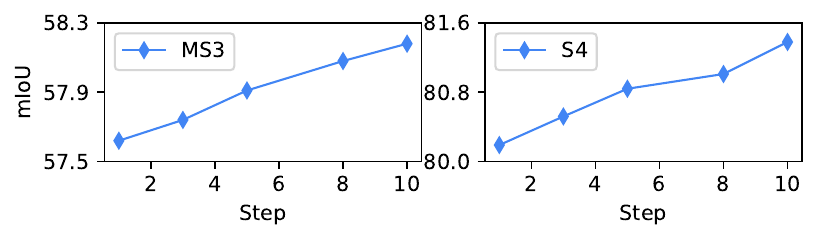}
\end{center}
% \vspace{-4.0mm}
\caption{\textbf{Performance with Different Denoising Steps.} The performance improves as the number of denoising steps increases, while we observe saturation after 10 steps.}
% \vspace{-4.0mm}
\label{fig:ddim_step}
\end{figure}

\noindent\textbf{Performance with Different Denoising Steps.}
The denoising step in diffusion models is usually pre-defined empirically. We set the denoising step in this paper following the conventional practice.
We thus evaluate the effect of the re-spaced inference denoising steps driven by the DDIM scheduler~\cite{song_DDIM_ICLR_2020}.
The change in testing performance for our model across the MS3 and S4 datasets with varying denoising steps is presented in Fig.~\ref{fig:ddim_step}.
Although the model is trained with 50 DDPM steps, employing 10 steps during inference is sufficient to achieve accurate results.
As expected, increasing the number of denoising steps leads to improved performance.
We observe that the elbow point of marginal returns given more denoising steps depends on the dataset but is always under 10 steps.
Hence, we determine that a denoising step value of 10 strikes an optimal trade-off between sampling efficiency and sample quality.

% It can be seen that improved performance is achieved with increased sampling steps.
% And when denoising step is set to 10, a saturated performance can be obtained. 
% Therefore, for the trade-off of sampling efficiency and sample quality, the value 10 is optimal for denoising step.
% larger
% along with the change in the 
% denoising steps is indeed beneficial for our task. 
% 
% To achieve trade-off between sampling efficiency and sample quality, we set denosing step as 10, and  in this paper.
% , the performance of our model shows a gradual increase.

\begin{figure}[t]
\begin{center}
\includegraphics[width=0.98\linewidth]{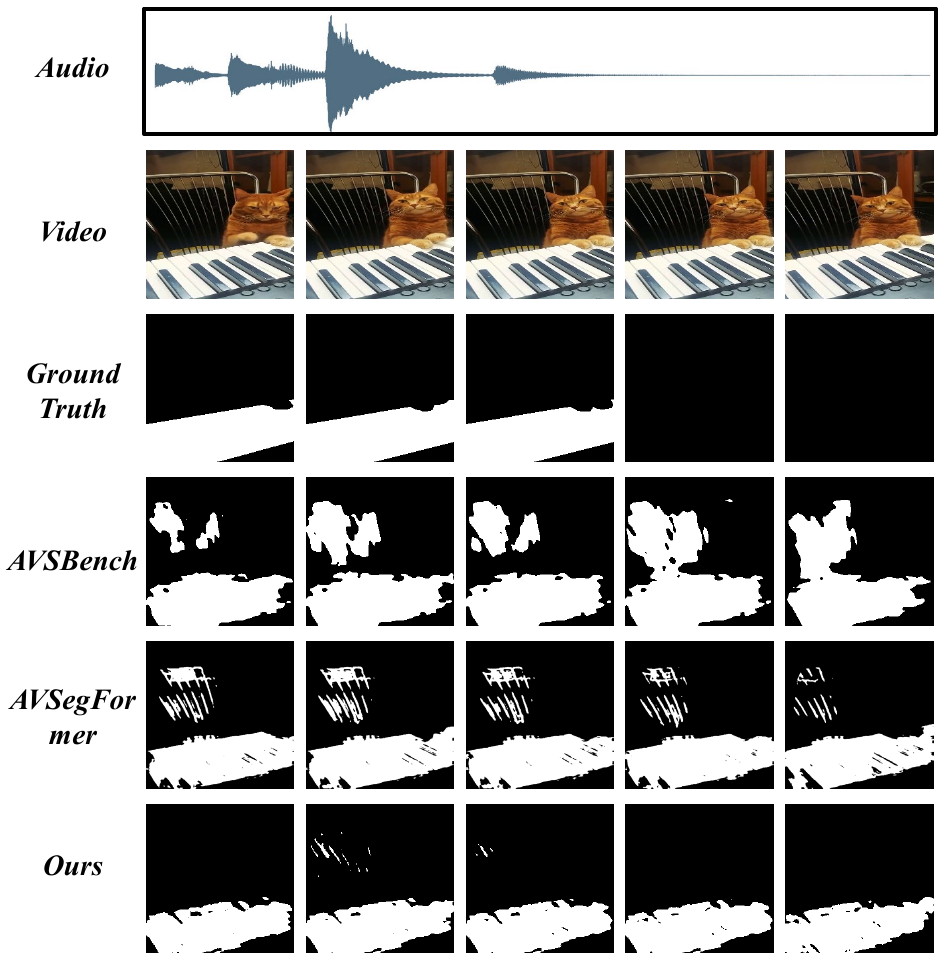}
\end{center}
% \vspace{-4.0mm}
\caption{\textbf{Failure case} on the fully-supervised MS3 setting.}
% \vspace{-4.0mm}
\label{fig:Failure}
\end{figure}

\noindent\textbf{Failure Case Analysis.}
\Fixtwo{We conduct a failure case analysis on our proposed method, AVSBench~\cite{zhou_AVSBench_ECCV_2022} and AVSegFormer~\cite{gao2024avsegformer}.}
In Fig.~\ref{fig:Failure}, it can be observed that our method, AVSbench, and \Fixtwo{AVSegFormer can not handle the absence of segmented objects resulting from sound interruptions.}
This limitation arises from the fact that neither our method nor AVSBench considered the \enquote{timing discontinuity} of the sound during the modeling process. 
Nevertheless, our proposed method is still able to achieve accurate sound source localization and then deliver high-quality segmentation results.
We believe that modeling from a temporal perspective, \ie~an audio-visual temporal correlation latent space, is one way to think about this problem.

\section{Conclusion}
% In this paper, we have 
We have proposed a conditional latent diffusion model with contrastive learning for audio-visual segmentation (AVS). 
We first define AVS as a guided binary segmentation task, where audio serves as the guidance for
% should be extensively explored to localize
segmenting the sound producer(s).
% in the visual data.
Based on the conditional setting, we have introduced a conditional latent diffusion model to
% As a conditional generation task, 
% we aim to 
maximize the conditional log-likelihood,
% which can be achieved with a conditional latent diffusion model, 
where the diffusion model is chosen to produce semantic correlated latent space.
Specifically, our latent diffusion model learns the conditional ground truth feature generation process, and the reverse diffusion process can then
% Especially, we have developed a latent diffusion model to perform the latent code of segmentation map generation conditioned on the audio-visual input, thus the reverse diffusion process can 
restore the ground-truth information during inference. 
Contrastive learning has been studied to further enhance the discriminativeness of the conditional variable, leading to mutual information maximization between the conditional variable and the final output. Quantitative and qualitative evaluations on the AVSBench dataset verify the effectiveness of our solution.

%%%%%%%%% REFERENCES
{
\bibliographystyle{ieeetr}
\bibliography{egbib}
}

%%%%%%%%% Biography

\end{document}